\documentclass[a4paper,11pt]{article}

\usepackage{geometry}
\usepackage[utf8]{inputenc}
\DeclareUnicodeCharacter{2212}{-}
\usepackage{amsmath}
\usepackage{microtype}
\usepackage{siunitx}
\usepackage{authblk}
\usepackage{lscape}
\usepackage{graphicx}
\usepackage{subcaption}
\usepackage[font={footnotesize},labelfont=bf]{caption}
\usepackage[super,numbers,sort&compress]{natbib}
\usepackage{booktabs}
\usepackage{blindtext}
\usepackage{booktabs,multirow,array}
\usepackage{tikz}
\usepackage{verbatim}
\usepackage{chemfig}
\usepackage{gensymb}
\usepackage{textcomp}
\usepackage{setspace}
\usepackage{multicol}
\usepackage{filecontents,pgfplots,pgfplotstable}
\usepackage{afterpage}
\usepackage{xtab,afterpage}
\usepackage{amssymb}
\usepackage{rotating}
\usepackage{natbib}
\usepackage{scrextend}
\usepackage{tcolorbox}
\usepackage[hidelinks]{hyperref}
\usepackage{cleveref}

\usepackage{enumitem}
\setlist[itemize]{leftmargin=*}
\usepackage{microtype}
\usepackage{float}
\usepackage{arxiv}

\begin{document}

\title{Pores for thought: Generative adversarial networks for stochastic reconstruction of 3D multi-phase electrode microstructures with periodic boundaries}

\author[1]{Andrea Gayon-Lombardo \thanks{a.gayon-lombardo17@imperial.ac.uk}\ \ }
\author[2]{Lukas Mosser}
\author[1]{Nigel P. Brandon \thanks{n.brandon@imperial.ac.uk}\ \ }
\author[3]{Samuel J. Cooper \thanks{samuel.cooper@imperial.ac.uk}\ \ }
\affil[1]{{\textit{\footnotesize Department of Earth Science and Engineering, Imperial College London, London SW7 2BP}}}
\affil[2]{{\textit{\footnotesize Earth Science Analytics ASA, Professor Olav Hanssens vei 7A, 4021 Stavanger, Norge}}}
\affil[3]{{\textit{\footnotesize Dyson School of Design Engineering, Imperial College London, London SW7 2DB}}}

\maketitle
\section*{Abstract}
{\footnotesize The generation of multiphase porous electrode microstructures is a critical step in the optimisation of electrochemical energy storage devices. This work implements a deep convolutional generative adversarial network (DC-GAN) for generating realistic n-phase microstructural data. The same network architecture is successfully applied to two very different three-phase microstructures: A lithium-ion battery cathode and a solid oxide fuel cell anode. A comparison between the real and synthetic data is performed in terms of the morphological properties (volume fraction, specific surface area, triple-phase boundary) and transport properties (relative diffusivity), as well as the two-point correlation function. The results show excellent agreement between datasets and they are also visually indistinguishable. 
By modifying the input to the generator, we show that it is possible to generate microstructure with periodic boundaries in all three directions. This has the potential to significantly reduce the simulated volume required to be considered “representative” and therefore massively reduce the computational cost of the electrochemical simulations necessary to predict the performance of a particular microstructure during optimisation. }


\section{Introduction}
 \begin{multicols}{2}
The geometrical properties of multiphase materials are of central importance to a wide variety of engineering disciplines. For example, the distribution of precious metal catalysts on porous supports; the structure of metallic phases and defects in high-performance alloys; the arrangement of sand, organic matter, and water in soil science; and the distribution of calcium, collagen and blood vessels in bone. \cite{Weyland2001, Mendez-Venegas2013, Fantazzini2003} In electrochemistry, whether we are considering batteries, fuel cells or supercapacitors, their electrodes are typically porous to maximise surface area but need to contain percolating paths for the transport of both electrons and ions, as well as maintaining sufficient mechanical integrity. \cite{Moussaoui2019, Cooper2017} Thus the microstructure of these electrodes significantly impacts their performance and their morphological optimisation is vital for developing the next generation of energy storage technologies. \cite{Moussaoui2018}\\

Recent improvements in 3D imaging techniques such as X-ray computed tomography (XCT) have allowed researchers to view the microstructure of porous materials at sufficient resolution to extract relevant metrics \cite{Holzer2011, Eastwood2014, Ni2016, Pietsch2017}. However, a variety of challenges remain, including how to extract the key metrics or “essence”   of an observed microstructural dataset such that synthetic volumes with equivalent properties can be generated, and how to modify specific attributes of this 

\end{multicols}
\twocolumn

microstructural data without compromising its overall resemblance to the real material. 

A wide variety of methods that consist of generating synthetic microstructure by numerical means have been developed to solve these challenges. \cite{Moussaoui2018} A statistical method for generating synthetic three-dimensional porous media based on distance correlation functions was introduced by Quiblier et al. \cite{Quiblier1984}. Following this work, Torquato et al. implemented a stochastic approach based on the n-point correlation functions for generating reconstructions of heterogeneous materials. \cite{Lu1990, Yeong1998, Yeong1998a, Manwart2000, Sheehan2001} Jiao et al. \cite{Jiao2007, Jiao2008} extended this method to present an isotropy-preserving algorithm to generate realisations of materials from their two-point correlation functions (TPCF).  Based on these previous works, the most widely used approach for reconstruction of microstructure implements statistical methods through the calculation of the TPCF. \cite{Sundararaghavan2014, Suzue2008, Hasanabadi2016, Hasanabadi2016a, Izadi2017, Mendez-Venegas2013, Hasanabadi2019}.\\

In the area of energy materials, interest has recently surged for generating synthetic microstructure in order to aid the design of optimised electrodes. The three-phase nature of most electrochemical materials adds an extra level of complexity to their generation compared to two-phase materials. Suzue et al. \cite{Suzue2008} implemented a TPCF from a two-dimensional phase map to reconstruct a three-dimensional microstructure of a porous composite anode. Baniassadi et al. \cite{Baniassadi2011} extended this method by adding a combined Monte Carlo simulation with a kinetic growth model to generate three-phase realisations of a SOFC electrode. Alternative algorithms for reconstruction of porous electrodes have been inspired by their experimental fabrication techniques. A stochastic algorithm based on the process of nucleation and grain growth was developed by Siddique et al. \cite{Siddique2010} for reconstructing a three-dimensional fuel cell catalyst layer. This process was later implemented by Siddique et al. \cite{Siddique2012} to reconstruct a three-dimensional three-phase LiFePO4 cathode. A common approach for generating synthetic microstructure of SOFC electrodes involves the random packing of initial spheres or “seeds”    followed by the expansion of such spheres to simulate the sintering process. \cite{Ali2008, Kenney2009, Cai2011, Bertei2012} Moussaoui et al. \cite{Moussaoui2018} implement a combined model based on sphere packing and truncated Gaussian random field to generate synthetic SOFC electrodes. Additional authors have implemented plurigaussian random fields    to model the three-phase microstructure of SOFC electrodes and establish correlations between the microstructure and model parameters \cite{LeLoch1997, Neumann2019, Neumann2019a, Moussaoui2019}.\\

In the area of Li-ion batteries some authors have implemented computational models to adhere a synthetic carbon-binder domain (CBD) (usually hard to image) into XCT three-dimensional images of the NMC/pore phases \cite{Usseglio-Viretta2018,Trembacki2018}. An analysis of transport properties such as tortuosity and effective electrical conductivity prove the effect of different CBD configurations in the electrode performance. Other authors have implemented physics-based simulations to predict the morphology of three-phase electrodes. Forouzan et al. \cite{Forouzan2016} developed a particle-based simulation that involves the superposition of CBD particles, to correlate the fabrication process of Li-ion electrodes with their respective microstructure. Srivastava et al.\cite{Srivastava2019} simulated the fabrication process Li-ion electrodes by controlling the adhesion of active material and CBD phases. Their physics-based dynamic simulations are able to predict with accuracy the effect of microstructure in transport properties.\\

\begin{figure*}[!h]
	\centering
	\includegraphics[clip, trim= .5cm 0cm 0cm 0cm, width=1\textwidth]{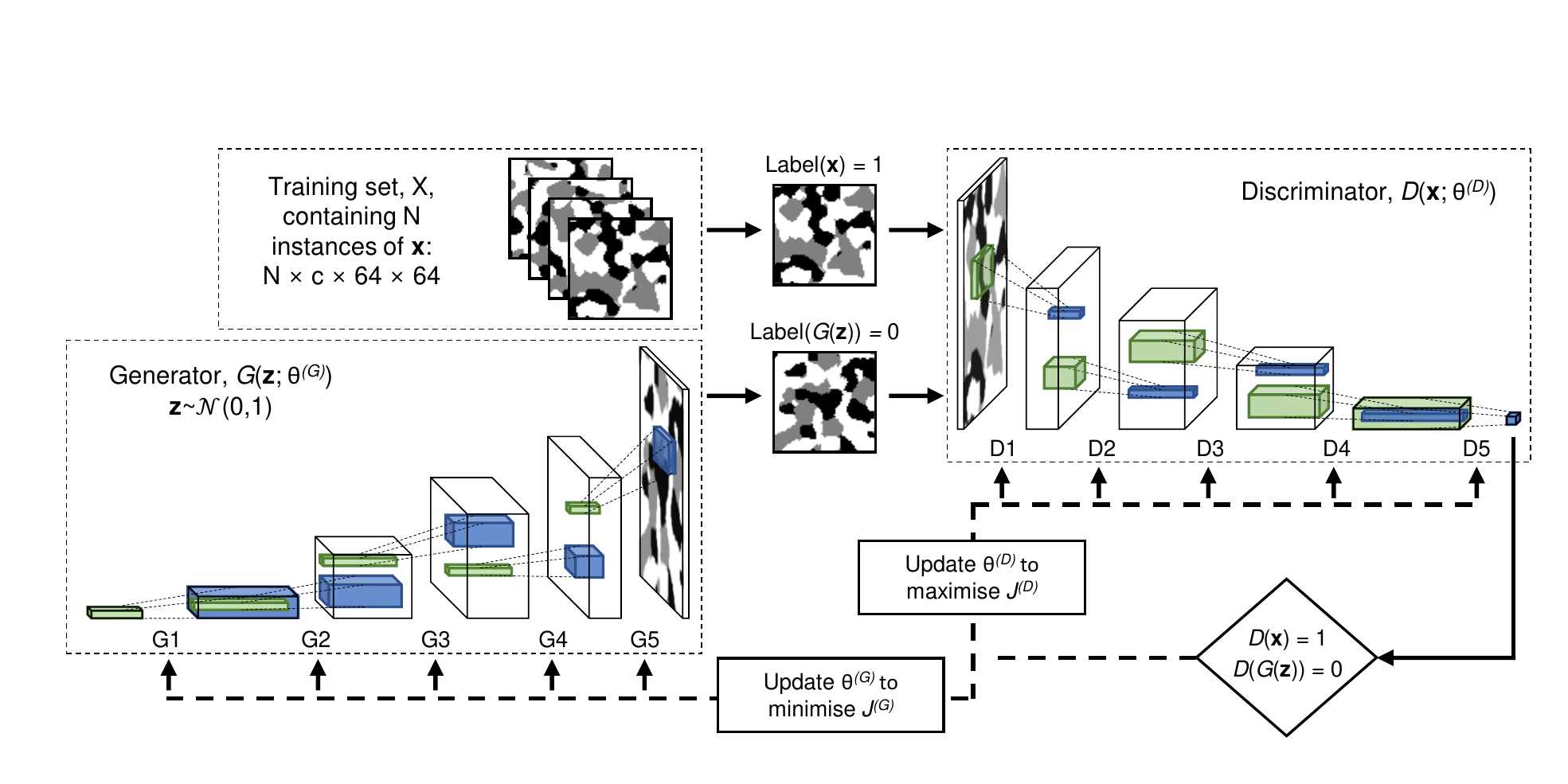}
	\caption[]
	{{\footnotesize \textbf{GAN architecture:} Schematic showing the architecture of the DC-GAN for 2D microstructural data. Generalisation to 3D samples is conceptually straightforward, but difficult to show as it requires the illustration of 4D tensors. In each layer, the green sub-volume shows a convolutional kernel at an arbitrary location in the volume and the blue sub-volume is the result of that convolution. In each case, the kernel is the same depth as the one non-spatial dimension, c, but must scan through the two spatial dimensions in order to build up the image in the following layer.}}    
	\label{Figure1_GAN}
\end{figure*}

Recent work by Mosser et al. \cite{Mosser2017, Mosser2018} introduces a deep learning approach for the stochastic generation of three-dimensional two-phase porous media. The authors implement a type of generative model called Generative Adversarial Networks (GANs)  \cite{Goodfellow2014} to reconstruct the three-dimensional microstructure of synthetic and natural granular microstructures. Li et al. \cite{Li2018} extended this work to enable the generation of optimised  sandstones, again using GANs. Compared to other common microstructure generation techniques, GANs are able to provide fast sampling of high-dimensional and intractable density functions without the need for an a priori model of the probability distribution function to be specified \cite{Mosser2018}. This work expands on the research of Mosser et al. \cite{Mosser2017, Mosser2018} and implements GANs for generating three-dimensional, three-phase microstructure for two types of electrode commonly used in electrochemical devices: a Li-ion battery cathode and an SOFC anode. A comparison between the statistical, morphological and transport properties of the generated images and the real tomographic data is performed. The two-point correlation function is further calculated for each of the three phases in the training and generated sets to investigate the long-range properties. Due to the fully convolutional nature of the GANs used, it is possible to generate arbitrarily large volumes of the electrodes based on the trained model. Lastly, by modifying the input of the generator, structures with periodic boundaries were generated. \\

Performing multiphysics simulations  on representative 3D volumes is necessary for microstructural optimisation, but it is typically very computationally expensive. This is compounded by the fact that the regions near the boundaries can show unrealistic behaviour due to the arbitrary choice of boundary condition. However, synthetic periodic microstructures (with all the correct morphological properties) enable the use of periodic boundary conditions in the simulation, which will significantly reduce the simulated volume necessary to be considered representative. This has the potential to greatly accelerate these simulations  and therefore the optimisation process as a whole.\\

The main contributions of this work are listed below:
 \begin{itemize}
 	\item The implementation of a GAN-based approach for the generation of multi-phase 3D microstructural data.
 	\item Application of this method to two types of commonly used three-phase electrodes, resulting trained generators that could be considered as ``virtual representations" of these microstructures. 
 	\item A statistical comparison of the microstructural properties between the real and generated microstructures, establishing the effectiveness of the approach.
 	\item The development of a method, based on the GAN approach, for generating periodic microstructures, and the implementation of a diffusion simulation on these volumes to illustrate the impact of periodic boundaries.
 	\item The extension of this approach for the generation of arbitrarily large, multiphase, periodic microstructures.
 \end{itemize}

\section{Generative Adversarial Networks}

Generative Adversarial Networks (GANs) are a type of deep generative model developed by Goodfellow et al. \cite{Goodfellow2014} which learn to implicitly represent the probability distribution function (pdf) of a given dataset (\textit{i.e.} $p_\mathrm{data}$)  . \cite{ Goodfellow2016} Since $p_\mathrm{data}$ is unknown, the result of the learning process is an estimate of the pdf called $p_\mathrm{model}$ from which a set of samples can be generated. Although GANs by design do not admit an explicit probability density, they learn a function that can sample from $p_\mathrm{model}$, which reasonably approximate those from the real dataset ($p_\mathrm{data}$).

The training process consists of a minimax game between two functions, the generator $G(\mathrm{\textbf{z}})$ and the discriminator $D(\mathrm{\textbf{x}})$. $G(\mathrm{\textbf{z}})$ maps an d-dimensional latent vector $\mathrm{\textbf{z}} \sim p_\mathrm{z}(\mathrm{\textbf{z}})\in \mathbb{R}^d$ to a point in the space of real data as $G(\mathrm{\textbf{z}};\theta^{(G)})$, while $D(\mathrm{\textbf{x}})$ represents the probability that $\mathrm{\textbf{x}}$ comes from $p_{\mathrm{data}}$.  \cite{Goodfellow2014} The aim of the training is to make the implicit density learned by $G(\mathrm{\textbf{z}})$ (\textit{i.e.} $p_{\mathrm{model}}$) to be close to the distribution of real data (\textit{i.e.} $p_{\mathrm{data}}$).  A more detailed introduction to GANs can be found in Section A of the supplementary material.\\

In this work, both the generator $G_{\theta^{(G)}}(\mathrm{\textbf{z}})$ and the discriminator $D_{\theta^{(D)}}(\mathrm{\textbf{x}})$ consist of deep convolutional neural networks. \cite{Lecun2015}   Each of these has a cost function to be optimised through stochastic gradient descent in a two-step training process. First, the discriminator is trained to maximise its loss function $J^{(D)}$:

\begin{multline}
J^{(D)} = \mathbb{E}_{\mathrm{\textbf{x}}\sim p_{\mathrm{data}}(\mathrm{\textbf{x}})} \left[\log \left(D(\mathrm{\textbf{x}})\right) \right] + \\ \mathbb{E}_{\mathrm{\textbf{z}}\sim p_{\mathrm{z}}(\mathrm{\textbf{z}})} \left[\log \left(1 - D\left(G(\mathrm{\textbf{z}})\right) \right) \right]
\end{multline}

This is trained as a standard binary cross-entropy cost in a classifier between the discriminator’s prediction and the real label. Subsequently, the generator is trained to minimise its loss function corresponding to minimising the log-probability of the discriminator being correct:

\begin{equation}
J^{(G)} = \mathbb{E}_{\mathrm{\textbf{z}}\sim p_{\mathrm{z}}(\mathrm{\textbf{z}})} \left[\log \left(1 - D\left(G(\mathrm{\textbf{z}})\right) \right) \right]
\end{equation}

These concepts are summarised in Figure \ref{Figure1_GAN}. Early in training, the discriminator significantly outperforms the generator, leading to a vanishing gradient in the generator. For this reason, in practice instead of minimising $ \log \left(1 - D\left(G(\mathrm{\textbf{z}})\right) \right)$, it is convenient to maximise the log-probability of the discriminator being mistaken, defined as $\log \left(D(G(\mathrm{\textbf{z}})) \right)$ \cite{Goodfellow2016}.\\

The solution to this optimisation problem is a Nash equilibrium  \cite{Goodfellow2016} where each of the players achieves a local minimum. Throughout the learning process, the generator learns to represent the probability distribution function $p_{\mathrm{model}}$ which is as close as possible to the distribution of the real data $p_{\mathrm{data}}(\mathrm{\textbf{x}})$.  At the Nash equilibrium, the samples of $\mathrm{\textbf{x}} = G(\mathrm{\textbf{z}}) \sim p_{\mathrm{model}} (\mathrm{\textbf{z}})$ are indistinguishable from the real samples $\mathrm{\textbf{x}} \sim p_{\mathrm{data}} (\mathrm{\textbf{x}})$   ,   thus $p_{\mathrm{model}} (\mathrm{\textbf{z}})=p_{\mathrm{data}} (\mathrm{\textbf{x}})$ and $D(\mathrm{\textbf{x}}) = \frac{1}{2}$ for all \textbf{x} since the discriminator can no longer distinguish between real and synthetic data.\\

\section{Microstructural image data}

\begin{figure*}[!h]
	\centering
	\includegraphics[clip, trim= 5cm 6cm 6.5cm 6cm, width=.9\textwidth]{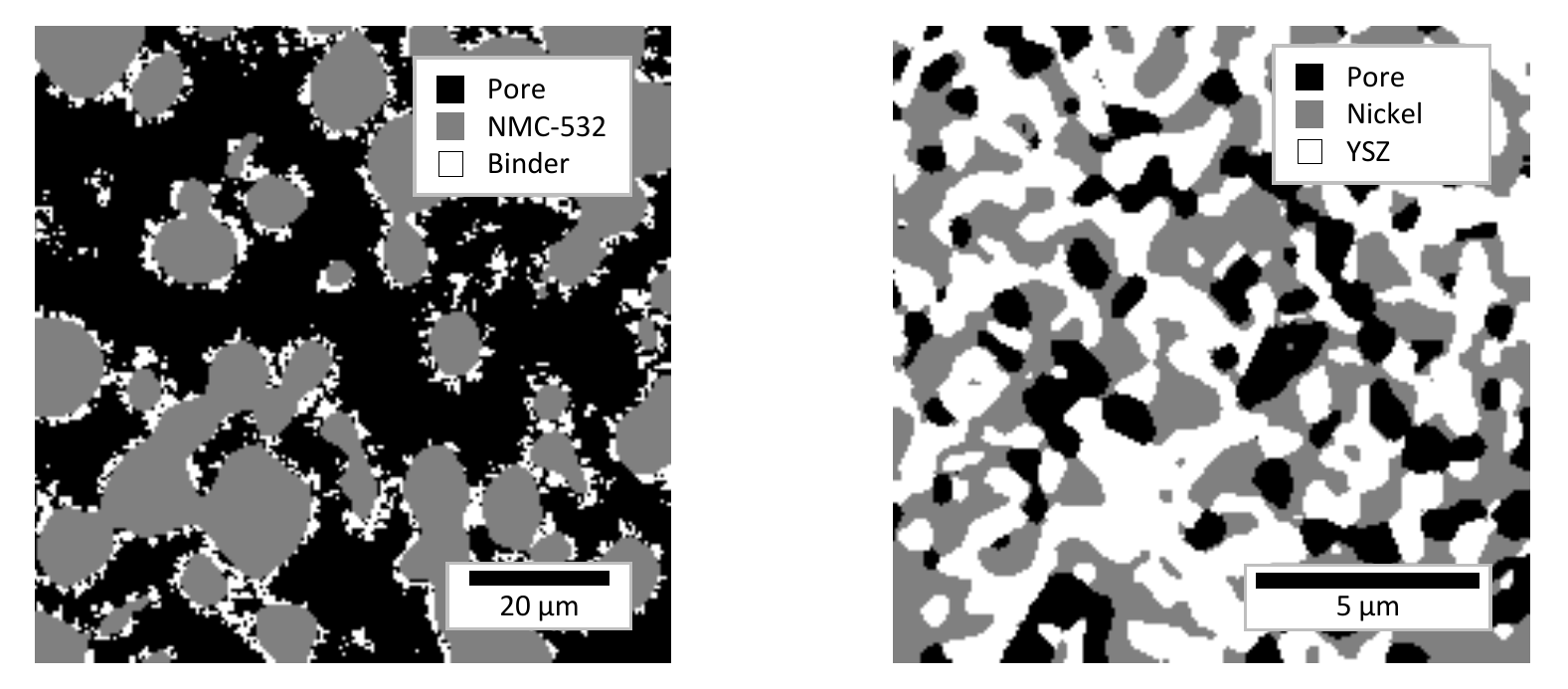}
	\caption[]
	{{\footnotesize \textbf{Original microstructures:} 2D images of the microstructures used as training sets in this work (L) Li-ion cathode, where the black phase represents pore, grey is NMC active material, and white is the organic binder; (R) SOFC anode, where the black phase corresponds to the pore, grey is Nickel and white is YSZ.}}    
	\label{Figure2_Microstructures}
\end{figure*}

\begin{table*}[]
\centering
	\caption{Key details about the two open-source, segmented, nano-tomography datasets}
	\begin{tabular}{c|c|c}
		\hline
		\textbf{Name}                            & \textbf{Li-ion  cathode} & \textbf{SOFC anode} \\
		\hline
		Materials  & NMC-532/Binder/Pore   & YSZ/Ni/Pore  \\
		Total volume of data / \si{\per\cubic\micro\meter} & $100.7 \times 100.3 \times 100.3$  & $124 \times 110 \times 8$ \\
		Number of voxels / voxels  & $253 \times 252 \times 252$  & $1900 \times 1697 \times 124$ \\
		Cubic voxel size / $\mathrm{nm^3}$  & $398^3$  & $65^3$ \\
		Number of training sub-volumes & 13,824 & 45,492 \\
		Reference    & \cite{Usseglio-Viretta2018} & \cite{Hsu2018} \\ 
		\hline
	\end{tabular}
	\label{Table_microstructure data}
\end{table*}

For both electrode materials considered in this study, open-source nano-tomography data was used which had already been segmented into each of their three respective phases (see Figure \ref{Figure2_Microstructures}). The first dataset is from a Li-ion battery cathode, comprising particles of a ceramic active material (nickel manganese cobalt oxide – NMC 532), a conductive organic binder (polymer with carbon black) and pores. This material is made by mixing the components in a solvent, thinly spreading this ink onto an aluminium foil and then drying. The second dataset is from an SOFC anode, comprising of a porous nickel/yttria stabilised zirconia (Ni-YSZ) cermet. This material is also made by mixing an ink, but this time it is deposited onto a ceramic substrate and then sintered at high temperature to bind the components together. Details of the sample preparation, imaging, reconstruction and segmentation approaches used can found in \cite{Usseglio-Viretta2018} for the cathode and \cite{Hsu2018} for the anode. The specifications of both datasets are shown in Table \ref{Table_microstructure data}.\\

More than ten-thousand overlapping sub-volumes were extracted from each image dataset (Table \ref{Table_microstructure data}), using a sampling function with a stride of 8 voxels.  The spatial dimensions of the cropped volumes were selected based on the average size of the largest structuring element. The sub-volume size was selected to guarantee that at least two structuring elements (\textit{i.e.} particular geometric shape that we attempt to extract from the XCT image) could fit in one sub-volume \cite{Blair1993}. In the case of the Li-ion cathode, this structuring element corresponds to the particle size. In the case of the SOFC anode, “particle size” is not easy to define once the sample is sintered, so the sub-volume was selected based on the stabilisation of the two-point correlation function (TPCF). A more detailed discussion of the representativeness of the sub-samples chosen for training is given in Section G of the supplementary information.\\

\section{Method}
This section outlines the architecture that constitutes the two neural networks in the GAN, as well as the methodology followed for training. It also describes the microstructural properties used to analyse the quality of the microstructural reconstruction when compared to the real datasets. The codes developed for training the GANs are open-source.\\

\subsection{Pre-treatment of the training set}
The image data used in this study is initially stored as 8-bit greyscale elements, where the value of each voxel (3D pixel) is used as a label to denote the material it contains. For example, as in the case of the anode in this study, black (0), grey (127), and white (255), encode pore, metal, and ceramic respectively.\\

In several previous studies where GANs were used to analyse materials, the samples in question had only two phases and as such, the materials information could be expressed by a single number representing the confidence it belongs to one particular phase. However, in cases where the material contains three or more phases, this approach can be problematic, as a voxel in which the generator had low confidence deciding between black or white may end up outputting grey (\textit{i.e.} a value halfway between black and white), which means something entirely different.\\

The solution comes from what is referred to as ``one-hot" encoding of the materials information. This means that an additional dimension is added to the dataset (\textit{i.e.} three spatial dimensions, plus one materials dimension), so an initially 3D cubic volume of n×n×n, is encoded to a 4D c×n×n×n volume, where c is the number of material phases present.  This materials dimension contains a ‘1’ in the element corresponding to the particular material at that location and a ‘0’ in all other elements (hence, “one-hot”). So, what was previous black, white, and grey, would now be encoded as [1,0,0], [0,1,0], and [0,0,1] respectively. This concept is illustrated in supplementary Figure S1 for a three-material sample, but it could trivially be extended to samples containing any number of materials. It is also easy to decode these 4D volumes back to 3D greyscale, even when there is uncertainty in the labelling, as the maximum value can simply be taken as the label.

\subsection{GAN Architecture and Training}
The GAN architecture implemented in this work is a volumetric version based on the deep convolutional GAN (DC-GAN) model proposed by Radford et al. \cite{Radford2015}. Both generator and discriminator are represented by fully convolutional neural networks. In particular, the convolutional nature of the generator allows it to be scalable, thus it can generate instances from the $p_\mathrm{model}$ larger than the instances in the original training set, which is useful.\\

\begin{table*}[!h]
	\caption{Dimensionality of each layer in the GAN architecture for each porous material (layers, dimensions, optimiser, input image size, number of training samples) See Figure \ref{Figure1_GAN}}
	\resizebox{\textwidth}{!}{
	\begin{tabular}{ccccccccc}
		\hline
		
		\multicolumn{1}{c|}{\textbf{Layer}} & \multicolumn{1}{c|}{\textbf{Function}} & \multicolumn{1}{c|}{\textbf{\begin{tabular}[c]{@{}c@{}}Input\\   channels\end{tabular}}} & \multicolumn{1}{c|}{\textbf{\begin{tabular}[c]{@{}c@{}}Output\\   channels\end{tabular}}} & \multicolumn{1}{c|}{\textbf{\begin{tabular}[c]{@{}c@{}}Spatial\\   Kernel\end{tabular}}} & \multicolumn{1}{c|}{\textbf{Stride}} & \multicolumn{1}{c|}{\textbf{Padding}} & \multicolumn{1}{c|}{\textbf{\begin{tabular}[c]{@{}c@{}}Batch\\   normalisation\end{tabular}}} & \textbf{\begin{tabular}[c]{@{}c@{}}Activation\\   function\end{tabular}} \\
		\hline
		Discriminator &  &  &  &  &  &  &  &  \\ \hline
		\multicolumn{1}{c|}{D1} & \multicolumn{1}{c|}{Conv3d} & \multicolumn{1}{c|}{3} & \multicolumn{1}{c|}{16} & \multicolumn{1}{c|}{$4 \times 4 \times 4$} & \multicolumn{1}{c|}{2} & \multicolumn{1}{c|}{1} & \multicolumn{1}{c|}{Yes} & LeakyReLU \\
		\multicolumn{1}{c|}{D2} & \multicolumn{1}{c|}{Conv3d} & \multicolumn{1}{c|}{16} & \multicolumn{1}{c|}{32} & \multicolumn{1}{c|}{$4 \times 4 \times 4$} & \multicolumn{1}{c|}{2} & \multicolumn{1}{c|}{1} & \multicolumn{1}{c|}{Yes} & LeakyReLU \\
		\multicolumn{1}{c|}{D3} & \multicolumn{1}{c|}{Conv3d} & \multicolumn{1}{c|}{32} & \multicolumn{1}{c|}{64} & \multicolumn{1}{c|}{$4 \times 4 \times 4$} & \multicolumn{1}{c|}{2} & \multicolumn{1}{c|}{1} & \multicolumn{1}{c|}{Yes} & LeakyReLU \\
		\multicolumn{1}{c|}{D4} & \multicolumn{1}{c|}{Conv3d} & \multicolumn{1}{c|}{64} & \multicolumn{1}{c|}{128} & \multicolumn{1}{c|}{$4 \times 4 \times 4$} & \multicolumn{1}{c|}{2} & \multicolumn{1}{c|}{1} & \multicolumn{1}{c|}{Yes} & LeakyReLU \\
		\multicolumn{1}{c|}{D5} & \multicolumn{1}{c|}{Conv3d} & \multicolumn{1}{c|}{128} & \multicolumn{1}{c|}{1} & \multicolumn{1}{c|}{$4 \times 4 \times 4$} & \multicolumn{1}{c|}{1} & \multicolumn{1}{c|}{0} & \multicolumn{1}{c|}{No} & Sigmoid \\ \hline
		Generator &  &  &  &  &  &  &  &  \\ \hline
		\multicolumn{1}{c|}{G1} & \multicolumn{1}{c|}{ConvTransposed3d} & \multicolumn{1}{c|}{100} & \multicolumn{1}{c|}{512} & \multicolumn{1}{c|}{$4 \times 4 \times 4$} & \multicolumn{1}{c|}{1} & \multicolumn{1}{c|}{0} & \multicolumn{1}{c|}{Yes} & ReLU \\
		\multicolumn{1}{c|}{G2} & \multicolumn{1}{c|}{ConvTransposed3d} & \multicolumn{1}{c|}{512} & \multicolumn{1}{c|}{256} & \multicolumn{1}{c|}{$4 \times 4 \times 4$} & \multicolumn{1}{c|}{2} & \multicolumn{1}{c|}{1} & \multicolumn{1}{c|}{Yes} & ReLU \\
		\multicolumn{1}{c|}{G3} & \multicolumn{1}{c|}{ConvTransposed3d} & \multicolumn{1}{c|}{256} & \multicolumn{1}{c|}{128} & \multicolumn{1}{c|}{$4 \times 4 \times 4$} & \multicolumn{1}{c|}{2} & \multicolumn{1}{c|}{1} & \multicolumn{1}{c|}{Yes} & ReLU \\
		\multicolumn{1}{c|}{G4} & \multicolumn{1}{c|}{ConvTransposed3d} & \multicolumn{1}{c|}{128} & \multicolumn{1}{c|}{64} & \multicolumn{1}{c|}{$4 \times 4 \times 4$} & \multicolumn{1}{c|}{2} & \multicolumn{1}{c|}{1} & \multicolumn{1}{c|}{Yes} & ReLU \\
		\multicolumn{1}{c|}{G5} & \multicolumn{1}{c|}{ConvTransposed3d} & \multicolumn{1}{c|}{64} & \multicolumn{1}{c|}{3} & \multicolumn{1}{c|}{$4 \times 4 \times 4$} & \multicolumn{1}{c|}{2} & \multicolumn{1}{c|}{1} & \multicolumn{1}{c|}{No} & Softmax \\ \hline
	\end{tabular}}
\label{Table_ Gan architecture}
\end{table*}

The discriminator is composed of five layers of convolutions, each followed by a batch normalisation. In all cases, the convolutions cover the full length of the materials dimension, \textit{c} but  the kernels within each layer are of smaller spatial dimension than the respective inputs to these layers . The first four layers apply a ``leaky" rectified linear unit (leaky ReLU) activation function and the last layer contains a sigmoid activation function that outputs a single scalar constrained between 0 and 1, as it is a binary classifier. This value represents an estimated probability of an input image to belong to the real dataset (output = 1) or to the generated sample (output = 0).\\

The generator is an approximate mirror of the discriminator, also composed of five layers, but this time transposed convolutions  \cite{Dumoulin2018} are used to expand the spatial dimensions in each step. Once again, each layer is followed by a batch normalization and all layers use ReLU as their activation function, except for the last layer which uses a Softmax  function, given by equation (\ref{eq_softmax})\\

\begin{multline}
\sigma(\mathrm{\textbf{x}})_i = \frac{e^{x_i}}{\sum_{j=1}^{K} e^{x_j}}\ \mathrm{for}\ i=1, \dots, K\ \mathrm{and}\ \\
\mathrm{\textbf{x}} = (x_1, \dots, x_K) \in \mathbb{R}^K
\label{eq_softmax}
\end{multline}

where $x_j$ represents the $j^\mathrm{th}$ element of the one-hot encoded vector $\mathrm{\textbf{x}}$ at the last convolutional layer.\\

It is well known that the hyperparmeters that define the architecture of the neural networks have significant impact on the quality of the results and the speed of training. In this work, although a formal hyperparameter optimisation was not performed due to computational expense, a total of 16 combinations between four hyperparmeters was performed. A statistical analysis between the real and generated microstructures was performed (as described in section \ref{results}), and the optimum architecture was chosen based on these results.\\

The generator requires a latent vector \textbf{z}  as its input in order to produce variety in its outputs.   In this study, the input latent vector   \textbf{z} is of length 100. Table \ref{Table_ Gan architecture} summarises all of the GAN’s layers configuration described above, as well as the size, stride and number of kernels applied between each layer, and the padding applied around the volume when calculating the convolutions. As will be discussed later in this paper, although zeros were initially used for padding, the study also explores the use of circular padding, which forces the microstructure to become periodic.\\

In theory, a Nash equilibrium is achieved after sufficient training; however, in practice this is not always the case. GANs have shown to present instability during training that can lead to mode collapse \cite{Goodfellow2016}. Mescheder et al. \cite{Mescheder2018} present an analysis of the stability of GAN training, concluding that instance noise and zero-centred gradient penalties lead to local convergence. Another proposed stabilisation mechanism, which was implemented effectively in this work, is called one-sided label smoothing \cite{Goodfellow2016}. Through this measure, the label 1 corresponding to real images is reduced by a constant $\varepsilon$, such that the new label has the value of $1 – \varepsilon$. For all cases in this work, $\varepsilon$ has a value of 0.1.\\

An additional source of instability during training is attributed to the fact that the discriminator learns faster than the generator, particularly at the early stages of training. To stabilise the alternating learning process, it is convenient to set a ratio of network optimisation for the generator and discriminator to $k:1$. In other words, the generator is updated k times while the discriminator is updated once. In this work $k$ has a value of 2.
In both cases (\textit{i.e.} cathode and anode data) stochastic gradient descent is implemented for learning using the ADAM optimiser \cite{Kingma2015}. The momentum constants are $\beta_1\ =\ 0.5,\ \beta_2\ =\ 0.999$ and the learning rate is $2 \times 10^{-5}$. All simulations are performed on a GPU (Nvidia TITAN Xp) and the training process is limited to 72 epochs (c. 48 h).\\

\subsection{Microstructural characterisation parameters}
The electrode microstructures can be characterised by a set of parameters calculated from the 3D volumes. These parameters include morphological properties, transport properties, and statistical correlation functions. To evaluate the ability of the trained model to accurately capture the pdf that describes the microstructure within the latent space, parameters were calculated from 100 instances of both the real and GAN generated data.

\subsubsection{Morphological properties}
Three morphological properties are considered in this work, each computed using the open-source software TauFactor \cite{Cooper2016}. These consist of the volume fractions and specific surface areas, as well as the triple-phase boundary (TPB) densities. More information about these parameters can be found in the supplementary information.

\subsubsection{Relative diffusivities}
The relative diffusivity, $D^{\mathrm{rel}}$, is a dimensionless measure of the ease with which diffusive transport occurs through a system held between Dirichlet boundaries applied to two parallel faces. In this study, it is calculated for each of the three material phases separately, as well as in each of the three principal directions in a cubic volume. It is directly related to the diffusive tortuosity factor of phase $i$, $\tau_i$, as can be seen from the following equation,

\begin{equation}
D^{\mathrm{rel}}_i = \frac{D^{\mathrm{eff}}_i}{D^0_i} = \frac{\phi_i}{\tau_i}
\label{eq_drel}
\end{equation}

where $\phi_i$ is the volume fraction of phase $i$, $D^0_i$  is the intrinsic diffusivity of the bulk material (arbitrarily set to unity), and $D^\mathrm{eff}_i$ is the calculated effective diffusivity given the morphology of the system. The tortuosity factors were obtained with the open-source software \textit{TauFactor} \cite{Cooper2016}, which models the steady-state diffusion problem using the finite difference method and an iterative solver.

\subsubsection{Two-point correlation function}
According to Lu et al. \cite{Lu1990} the morphology of heterogeneous media can be fully characterised by specifying one of the various statistical correlation functions. One of such correlations is the \textit{n}-point probability function $S_n (\mathrm{\textbf{x}}^{n})$, defined as the probability of finding n points with positions $\mathrm{\textbf{x}}_n$ in the same phase \cite{Lu1990, Yeong1998, Yeong1998a}. Based on this, the so-called two-point correlation function (TPCF), $S_2 (r)$, allows the first and second-order moments of a microstructure to be characterised \cite{Yeong1998a, Mosser2018}. Assuming stationarity (\textit{i.e.} the mean and variance have stabilised), the TPCF is defined as the non-centred covariance, which is the probability $P$ that two points $x_1  = x$ and $x_2  = x + r$ separated by a distance $r$ belong to the same phase $p_i$,

\begin{equation}
S_2(r) = P(x\ \in\ p_i,\ x+r\ \in\ p_i)\ \mathrm{for}\ x,r\ \in\ \mathbb{R}^d
\label{eq_TPCF}
\end{equation}

At  the origin, $S_2(0)$ is equal to the phase volume fraction $\phi_i$. The function $S_2(r)$ stabilised at the value of $\phi_i^2$ as the distance, $r$, tends to infinity.This function is not only valuable for analysing the anisotropy of the microstructure, but also to account for the representativeness in terms of volume fraction of sub-volumes taken from the same microstructure sample. In this work, the TPCF of the three phases is calculated along the three Cartesian axes.

\section{Results} \label{results}
The two GANs implemented in this work, one for each microstructure, were trained for a maximum of 72 epochs (\textit{i.e.} 72 complete iterations of the training set). The stopping criteria was established through the manual inspection of the morphological properties every two epochs. Figures S9 and S10 of the supplementary data shows the visual reconstruction of both microstructures, beginning with Gaussian noise at epoch 0, and ending with a visually equivalent microstructure at epoch 50. The image generation improves with the number of iterations; however, as pointed out by Mosser et al. \cite{Mosser2017}, this improvement cannot be observed directly from the loss function of the generator and so the morphological parameters described above are used instead.\\

\begin{figure*}[h]
	\centering
	\includegraphics[clip, trim= 2cm .5cm 4cm 0cm, width=.95\textwidth]{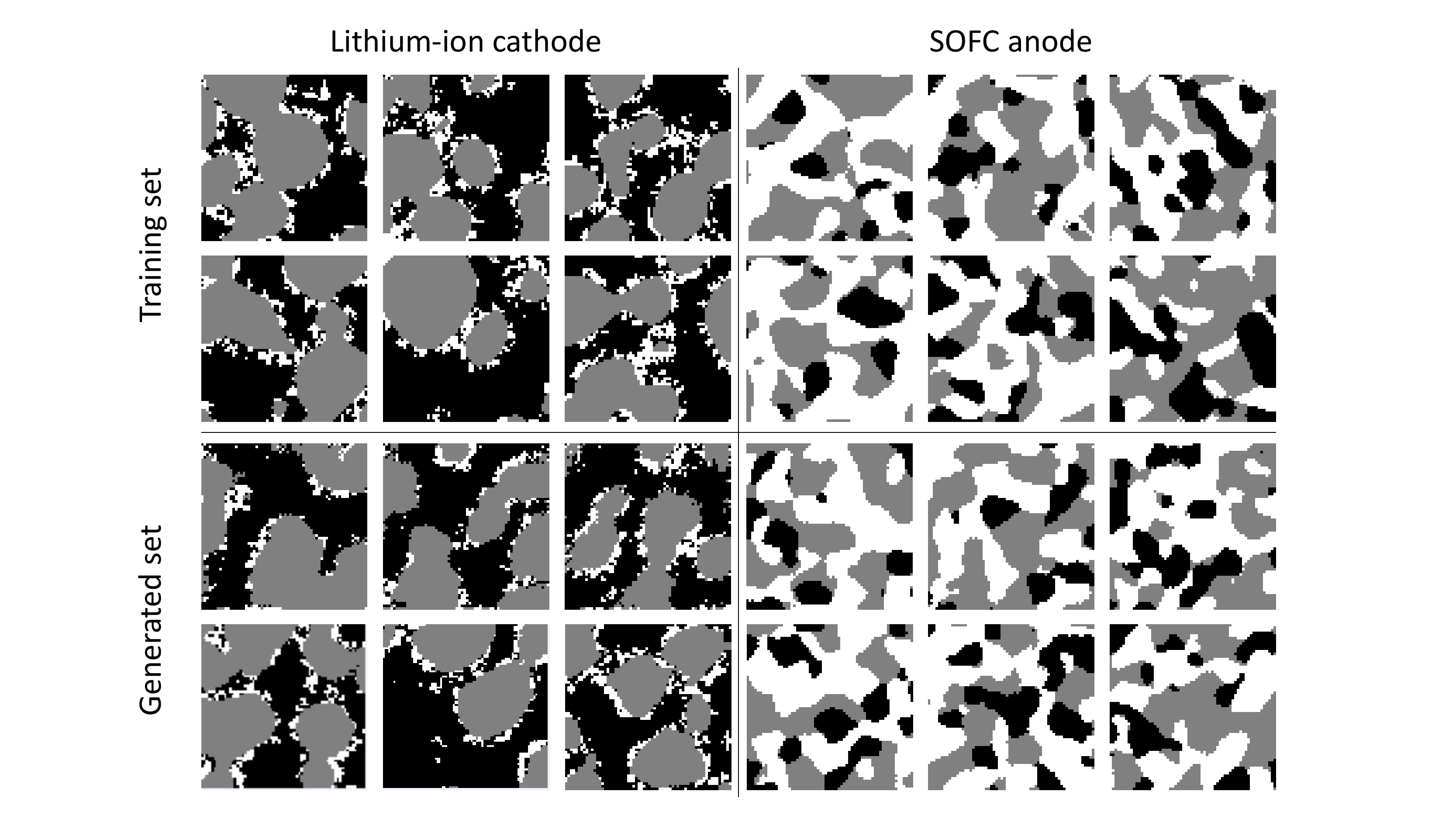}
	\caption[]
	{{\footnotesize \textbf{Generated vs real:} Images from both the cathode and anode samples, illustrating the excellent qualitative correspondence between the real and generated microstructures.}}    
	\label{Figure3_generated_real}
\end{figure*}

Before introducing the quantitative comparison of the real and synthetic microstructural data, the excellent qualitative agreement between them can be seen in Figure \ref{Figure3_generated_real}, which shows six instances of both real and synthetic data from both the anode and cathode samples. Each slice consists of $64^2$ voxels and is obtained from a $64^3$ generated volumetric image. This qualitative analysis consisted of visually comparing some key features of the data. In the case of the Li-ion cathode, the structuring element in the microstructure (\textit{i.e.} the NMC particles – grey phase) shows round borders surrounded by a thin layer of binder (white phase), and the phases seem distributed in the same proportion in the real and generated images. In the case of the SOFC, no structuring element is clearly defined; however, the shapes of the white and grey phases of the generated set show the typical shape of the real data which is particular for the sintering process in the experimental generation of these materials \cite{Hsu2018}.\\

In order to draw a comparison between the training and generated microstructures, 100 instance of each were randomly selected ($64 \times 64 \times 64$ voxels). The comparison results are presented in the following sections. Figure S9 and S10 show 2D slices through generated volumes containing the max value in the phase dimension (rather than the label), which indicates the confidence with which it was labelled.

\subsection{Lithium-ion cathode results}
\begin{figure*}[]
	\centering
	\includegraphics[clip, trim= 0.5cm 5cm 0cm 6cm, width=1\textwidth]{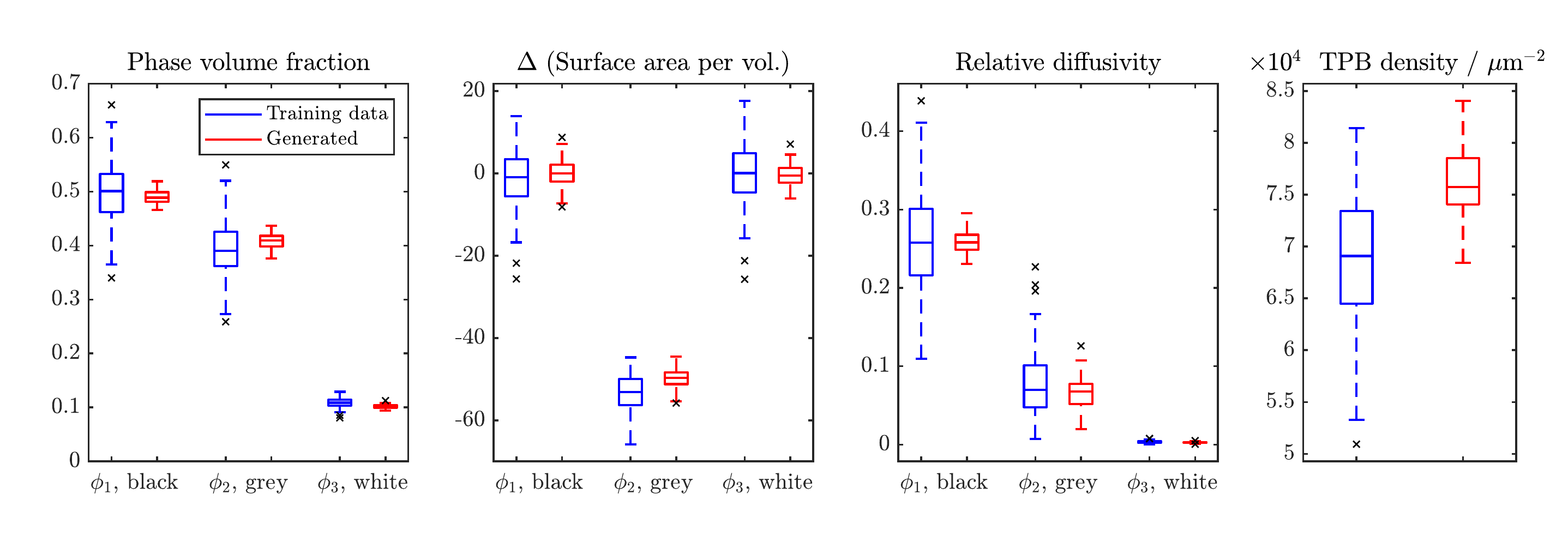}
	\caption[]
	{{\footnotesize \textbf{Characterisation properties Li-ion:} Microstructural characterisation properties for 100 training samples (blue) and 100 generated realisations (red) for the Li-ion cathode. Black crosses show outliers. In all cases, greater diversity is observed in the training set than the generated samples.}}    
	\label{Figure4_Characteristics_Liion}
\end{figure*}

The results of the calculation of the microstructural characterisation parameters (\textit{i.e.} phase volume fraction, specific surface area and TPB) for the three phases are presented in Figure \ref{Figure4_Characteristics_Liion}. For ease of comparison in a single figure, the results of the specific surface area analysis are presented in terms of the percentage deviation from the maximum mean area value among the three phases, $\Delta(SSA)$. In the case of the cathode, the maximum mean area, $A_{\mathrm{max,mean}}$, corresponds to the mean area of the white phase (binder) of the training set ($A_{\mathrm{max,mean}}=$ \SI{0.72}{\per\micro\meter}) and all other areas, $A_i$, are normalised against this.\\

\begin{equation}
\Delta(SSA) = \frac{A_i - A_{\mathrm{max,mean}}}{A_{\mathrm{max,mean}}}
\label{eq_DeltaSSA}
\end{equation}

\begin{table*}[h]
	\caption{Results of volume fractions, specific surface areas, triple-phase boundary densities and relative diffusivities calculated from the real and generate datasets. The black phase corresponds to the pores, white phase corresponds to the binder and grey phase corresponds to the NMC-532.}
	\centering
	\begin{tabular}{c|c|c|c|c|c}
		\hline
		\textbf{Dataset} & \textbf{Phase} & \textbf{Volume fraction} & \textbf{\begin{tabular}[c]{@{}c@{}}Specific surface\\   area / \si{\per\micro\meter} \end{tabular}} & \textbf{\begin{tabular}[c]{@{}c@{}}TPB \\ density / \si{\per\square\micro\meter} \end{tabular}} & \textbf{$D^{\mathrm{rel}}$} \\ \hline
		\multirow{3}{*}{Real} & Black & 0.50 ± 0.06 & 0.71 ± 0.06 & \multirow{3}{*}{0.43 ± 0.04} & 0.26 ± 0.07 \\
		 & Grey & 0.39 ± 0.06 & 0.34 ± 0.03 &  & 0.08 ± 0.04 \\
		& White & 0.11 ± 0.01 & 0.72 ± 0.06 &  & 0.003 ± 0.001 \\ \hline
		\multirow{3}{*}{Generated} & Black & 0.49 ± 0.01 & 0.72 ± 0.02 & \multirow{3}{*}{0.48 ± 0.02} & 0.26 ± 0.02 \\
		& Grey & 0.41 ± 0.01 & 0.36 ± 0.02 &  & 0.07 ± 0.02 \\
		& White & 0.102 ± 0.003 & 0.72 ± 0.02 &  & 0.003 ± 0.001 \\ \hline
	\end{tabular}
\label{Table_li-ion results}
\end{table*}

The phase volume fraction, specific surface area and relative diffusivity show good agreement between the real and the synthetic data, particularly in the mean values of both distributions. These mean values and standard deviations are reported in Table \ref{Table_li-ion results}. The distribution of relative diffusivity in the white phase is very close to zero due to the low volume fraction and resulting low percolation of this phase. For the TPB density, the mean of the generated set is nearly 10\% greater than that of the training data; however, nearly all of the values for the synthetic data do still fall within the same range as the training set.\\

From Figure \ref{Figure4_Characteristics_Liion} it is clear that the synthetic realisations show a smaller variance in all of the calculated microstructural properties compared to the real datasets. This is also shown in Figure S2 where the relative diffusivity is plotted against the volume fraction of the three phases. \\

\begin{figure}[]
	\centering
	\includegraphics[clip, trim= 8.2cm 4cm 9cm 5cm, width=0.46\textwidth]{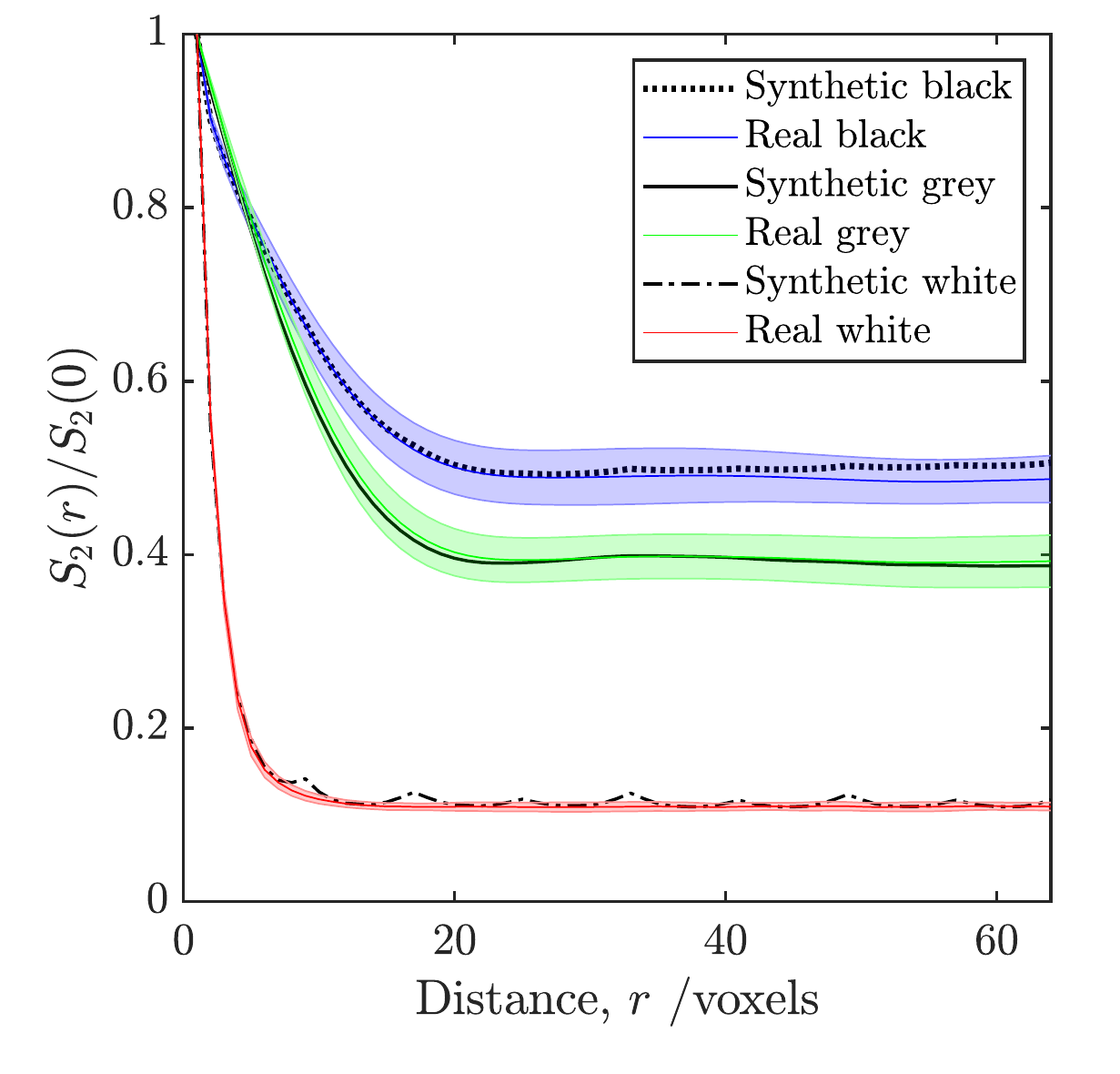}
	\caption[]
	{{\footnotesize \textbf{TPCF Li-ion cathode} Averaged TPCF for the three phases present the Li-ion cathode. The averaged values are obtained from 100 training samples and 100 synthetic realisations generated with the GAN model. The coloured band represents the standard deviation of the metric from the real data at each value or $r$.}}    
	\label{Figure5_TPCF_Liion}
\end{figure}

The averaged $S_2(\mathrm{\textbf{r}})/S_2 (0)$ along three directions is shown in Figure \ref{Figure5_TPCF_Liion} for each of the three phases present in a Li-ion cathode. Since $S_2 (0)$ represents the volume fraction $\phi_i$ of each phase, $S_2(\mathrm{\textbf{r}})/S_2 (0)$ is a normalisation of the TPCF that ranges between 0 and 1. In this expression, $S_2(\mathrm{\textbf{r}})/S_2 (0)$ stabilised at the value of $\phi_i$. In all cases, the average values of $S_2(\mathrm{\textbf{r}})/S_2 (0)$ of the synthetic realisations follow the same trend as the training data. The black phase shows a near exponential decay, the grey phase presents a small hole effect  and the white phase shows an exponential decay. A hole effect is present in a two-point correlation function when the decay is non-monotonic and presents peaks and valleys. This property indicates a form of pseudo-periodicity and in most cases is linked to anisotropy \cite{Journel1982, Pyrcz2003}. For the black and grey phases, the $S_2(\mathrm{\textbf{r}})/S_2 (0)$ values of the generated images show a slight deviation from the training set, however this value falls within the standard deviation of the real data.\\

\begin{table*}[!h]
	\caption{Results of volume fractions, specific surface areas, triple-phase boundary densities and relative diffusivities calculated from the real and generate datasets. The black phase corresponds to the pores, white phase corresponds to the ceramic (\textit{i.e.} YSZ) and grey phase corresponds to the metal (\textit{i.e.} Ni).}
	\centering
		\begin{tabular}{c|c|c|c|c|c}
			\hline
			\textbf{Dataset} & \textbf{Phase} & \textbf{Volume fraction} & \textbf{\begin{tabular}[c]{@{}c@{}}Specific surface\\   area / \si{\per\micro\meter} \end{tabular}} & \textbf{\begin{tabular}[c]{@{}c@{}}TPB \\ density / \si{\per\square\micro\meter} \end{tabular}} & \textbf{$D^{\mathrm{rel}}$} \\ \hline
			\multirow{3}{*}{Real} & Black & 0.21 ± 0.04 & 2.15 ± 0.17 & \multirow{3}{*}{8.10 ± 0.97} & 0.01 ± 0.03 \\
			& Grey & 0.34 ± 0.02 & 3.66 ± 0.21 &  & 0.10 ± 0.02 \\
			& White & 0.45 ± 0.03 & 3.93 ± 0.16 &  & 0.19 ± 0.03 \\ \hline
			\multirow{3}{*}{Generated} & Black & 0.21 ± 0.01 & 2.24 ± 0.06 & \multirow{3}{*}{7.37 ± 0.32} & 0.01 ± 0.01 \\
			& Grey & 0.35 ± 0.01 & 3.71 ± 0.08 &  & 0.10 ± 0.01 \\
			& White & 0.45 ± 0.01 & 3.96 ± 0.06 &  & 0.20 ± 0.01 \\ \hline
	\end{tabular}
	\label{Table_SOFC results}
\end{table*}
\begin{figure*}[!h]
	\centering
	\includegraphics[clip, trim= 0.5cm 5cm 0cm 6cm, width=1\textwidth]{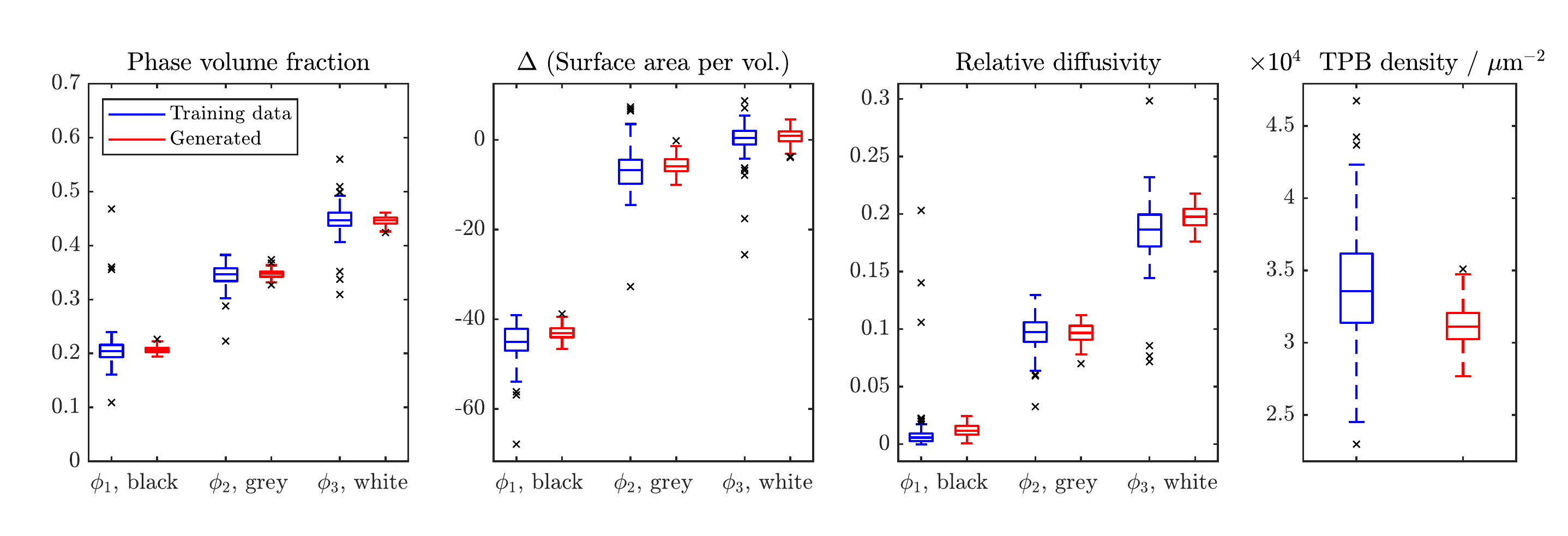}
	\caption[]
	{{\footnotesize \textbf{Characterisation properties SOFC:} Microstructural characterisation properties for 100 training samples and 100 generated realisations for the SOFC anode}}    
	\label{Figure6_Characteristics_SOFC}
\end{figure*}

\subsection{SOFC anode results}

Figure \ref{Figure6_Characteristics_SOFC} presents the results of the SOFC anode microstructural characterisation parameters calculated for the training data and for the synthetic realisations generated with the GAN model. The $\Delta(SSA)$ was calculated using the same approach as described in the previous section, but in the case of the anode, the maximum mean area was the mean area of the white phase (YSZ) of the training set ($A_{\mathrm{max,mean}}=$ \SI{3.98}{\per\micro\meter})\\

\begin{figure}[]
	\centering
	\includegraphics[clip, trim= 8.2cm 4cm 9cm 5cm, width=0.46\textwidth]{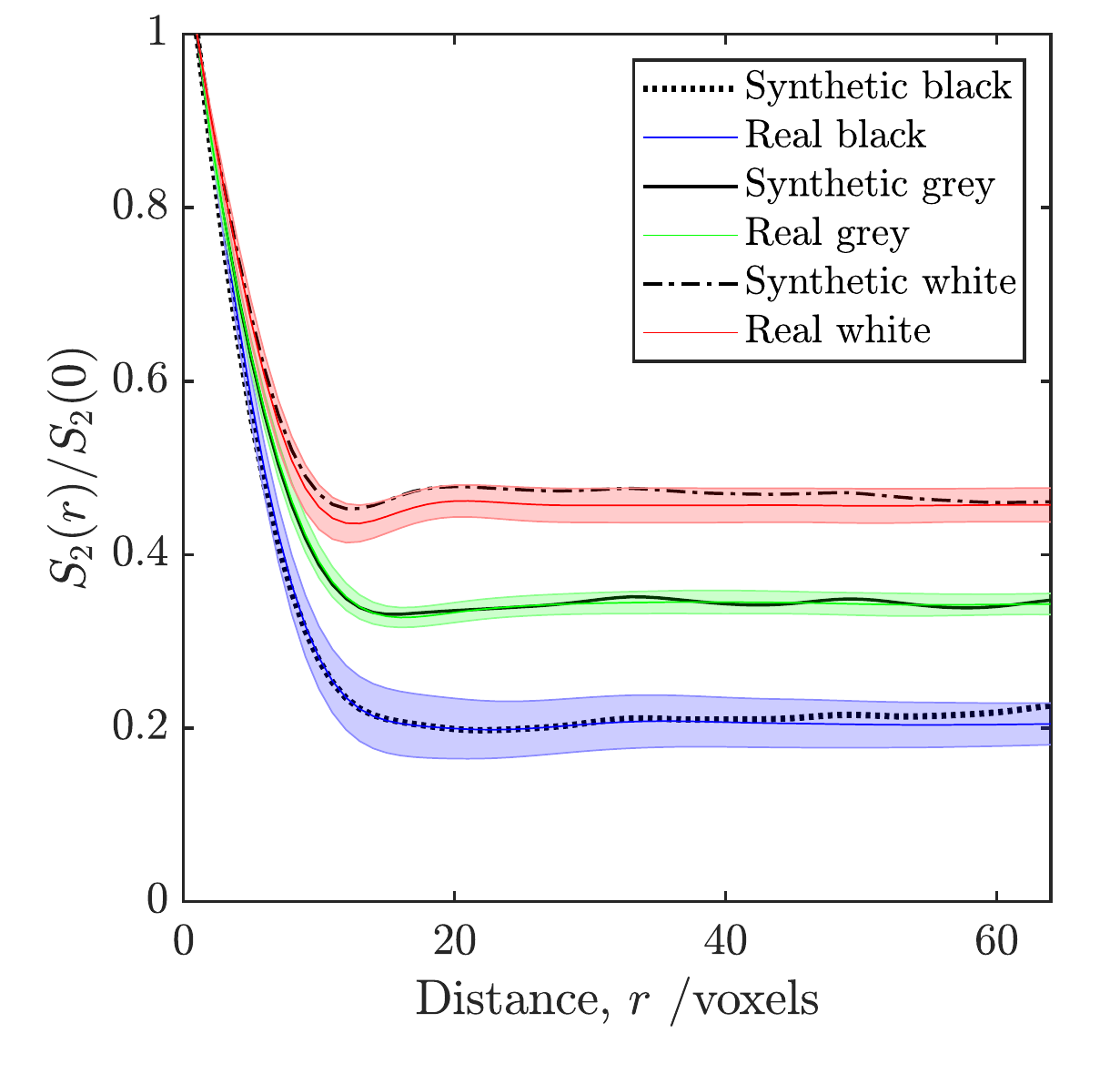}
	\caption[]
	{{\footnotesize \textbf{TPCF SOFC anode:} Averaged TPCF for the three phases present the SOFC anode. The averaged values are obtained from 100 training samples and 100 synthetic realisations generated with the GAN model}}    
	\label{Figure7_TPCF_SOFC}
\end{figure}

The results in Figure \ref{Figure6_Characteristics_SOFC} show a comparable mean and distributions for the morphological properties calculated, as well as for the effective diffusivity of the training images and the synthetic realisations. Once again, the synthetic images show lower variance in the calculated properties than the training set. These mean values and standard deviations are reported in Table \ref{Table_SOFC results}.\\

Once again, the difference in the diversity of synthetic images with respect to the training set can be clearly seen in supplementary Figure S3 where the effective diffusivity averaged over the three directions for each phase is plotted against its respective volume fraction.\\

\begin{figure*}[!h]
	\centering
	\includegraphics[clip, trim= 1cm 7cm 1cm 1cm, width=1\textwidth]{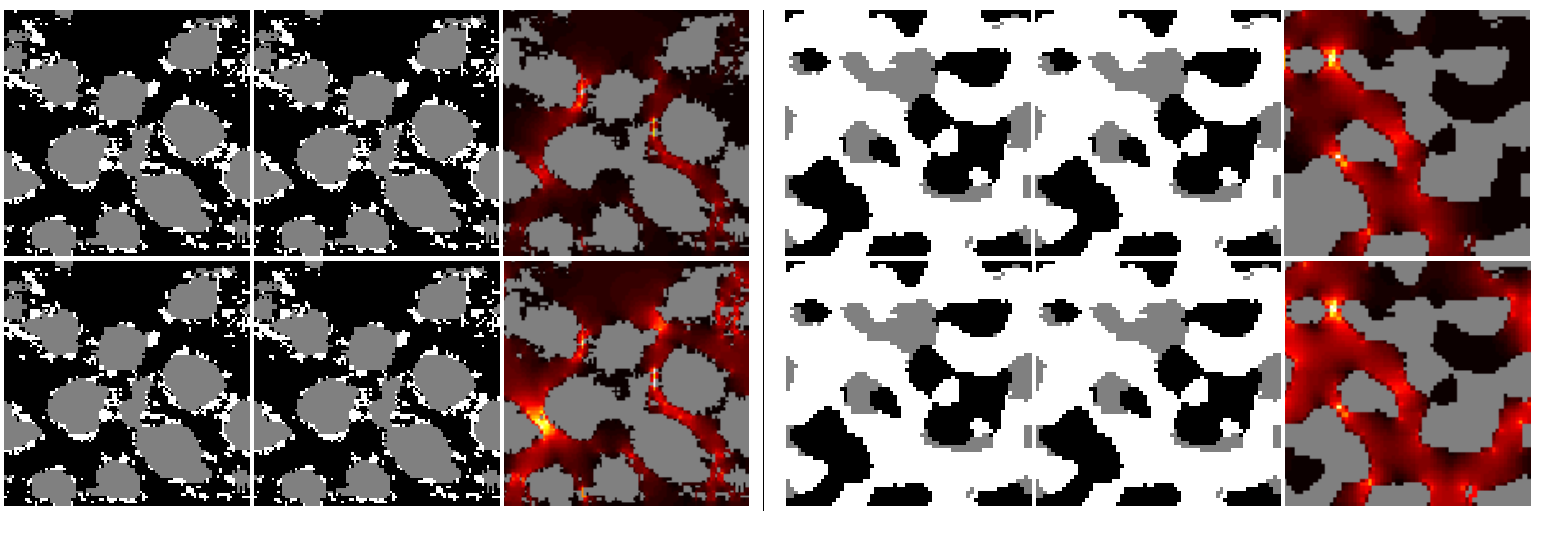}
	\caption[]
	{{\footnotesize \textbf{Periodic boundaries:} For both the Li-ion cathode (L) and the SOFC anode (R), periodic microstructures were generated by slightly changing the input to the generator. For each electrode, four instances are shown making the periodicity easier to observe. Also shown are local scalar flux maps generated from steady-state diffusion simulations in TauFactor with either mirror (top) or periodic (bottom) boundary conditions implemented on the vertical edges.}}    
	\label{Figure8_Periodic_boundaries}
\end{figure*}

The TPCF was calculated for the three phases along the three Cartesian directions. Figure \ref{Figure7_TPCF_SOFC} shows the value of $S_2(\mathrm{\textbf{r}})/S_2 (0)$ for the three phases present in the SOFC anode. The averaged results show an exponential decay in the black phase, a small hole effect \cite{Pyrcz2003} in the grey phase and a pronounced hole effect in the white phase.\\

\subsection{Periodic boundaries}

Once the generator parameters have been trained, the generator can be used to create periodic microstructures of arbitrary size. This is simply achieved by applying circular spatial padding to the first transposed convolutional layer of the generator (although other approaches are possible). Figure \ref{Figure8_Periodic_boundaries} shows generated periodic microstructures for both the cathode and anode, arranged in an array to make their periodic nature easier to see.  Additionally, local scalar flux maps resulting from steady-state diffusion simulations in \textit{TauFactor} \cite{Cooper2016} are shown for each microstructure. In both cases, the upper flux map shows the results of the simulation with mirror (\textit{i.e.} zero flux) boundaries on the vertical edges, and the lower one shows the results of the simulation with periodic boundaries on the vertical edges. Comparing the results from the two boundary conditions, it is clear that using periodic boundaries opens up more paths that enable a larger flux due to the continuity of transport at the edges. Furthermore, this means that the flow effectively does not experience any edges in the horizontal direction, which means that, unlike the mirror boundary case, there are no unrealistic regions of the volume due to edge effects.\\

\section{Discussion}
This work presents a technique for generating synthetic three-phase, three-dimensional microstructure images through the use of deep convolutional generative adversarial networks. The main contributions of this methodology are mentioned as follows: 
The results from comparing the morphological metrics, relative diffusivities and two-point correlation functions all show excellent agreement between the real and generated microstructure. One of the properties that is different from the averaged value in both cases is the TPB density. Nevertheless, its value falls within the confidence interval of the real dataset. This comparison demonstrates that the stochastic reconstruction developed in this work is as accurate as the state-of-the-art reconstruction methods reported in the introduction of this article. One variation from previous methodologies is that GANs do not require additional physical information from the microstructure as input data. The methodology was developed to approximate the probability distribution function of a real dataset, so it learns to approximate the voxel-wise distribution of phases, instead of directly inputting physical parameters, which is significant; although inputting physical parameters in additional may be beneficial \cite{Paganini2018}.\\

Despite the accurate results obtained in terms of microstructural properties, a number of questions still need to be addressed. One of which involves the diversity in terms of properties of the generated data. Particularly in the case of the Li-ion cathode microstructure, the generated samples present less variation than the training set.  This issue was already encountered by Mosser et al. \cite{Mosser2018} and low variation in the generated samples is a much discussed issue in the GAN literature. The typical explanation for this is based on the original formulation of the GAN objective function, which is set to represent unimodal distributions, even when the training set is multimodal \cite{Goodfellow2016, Goodfellow2014, Mosser2018}. This behaviour is known as “mode collapse” and is observed as a low variability in the generated images. A visual inspection of the generated images as well as the accuracy in the calculated microstructural properties do not provide a sufficient metric to guarantee the inexistence of mode collapse or memorisation of the training set.\\

Figures \ref{Figure4_Characteristics_Liion} and \ref{Figure6_Characteristics_SOFC} show some degree of mode collapse given by the small variance in the calculated properties  of the generated data. Nevertheless, further analysis of the diversity of the generated samples is required to evaluate the existence of mode collapse based on the number of unique samples that can be generated \cite{Heusel2017, Arora2018}.  Following the work of Radford et al. \cite{Radford2015}, an interpolation between two points in the latent space is performed to test the absence of memorisation in the generator. The results shown in Figure S8 present a smooth transformation of the generated data as the latent vectors is progressed along a straight path  . This indicates that the generator is not memorising the training data but has learned the meaningful features of the microstructure.\\

The presence of mode collapse and vanishing gradient remain the two main issues with the implementation of GANs. As pointed out by  \cite{Wang2019}, these problems are not necessarily related to the architecture of GANs, but potentially to the original configuration of the loss function. This work implements a DC-GAN architecture with the standard loss function; however, recent improvements of GANs have focused on reconfiguring the loss function to enable a more stable training and more variability in the output. Some of these include WGAN (and WGAN-GP) based on the Wasserstein or Earthmover distance \cite{Arjovsky2017, Gulrajani2017}, LSGAN which uses least squares and the Pearson divergence \cite{Mao2017}, SN-GAN that implement a spectral normalisation \cite{Yoshida2017}, among others \cite{Wang2019}. Therefore, an improvement of the GAN loss function is suggested as future work in order to solve the problems related to low variability (i.e. slight mode collapse) and training stability.\\

The applicability of GANs can be extended to transfer the learned weights of the generator (\textit{i.e.} $G_\theta (\mathrm{\textbf{z}})$) into a) generating larger samples of the same microstructure, b) generating microstructure with periodic boundaries, c) performing an optimisation of the microstructure according to a certain macroscopic property based on the latent space z   . As such, $G_\theta$ can be thought of as a powerful ``virtual representation" of the real microstructure and it interested to note that the total size of the trained parameters, $\theta^{(G)}$ is just 55 MB.\\

The minimum generated samples are the same size as the training data sub-volumes (\textit{i.e.} $64^3$ for both cases analysed in this work), but can be increased to any arbitrarily large size by increasing the size of the input \textbf{z}. Although the training process of the DC-GAN is computationally expensive, once a trained generator is obtained, it can produce image data inexpensively. The relation between computation time and generated image size is shown in Figure S7.\\

The ability of the DC-GAN to generate periodic structures has potentially profound consequences for the simulation of electrochemical processes at the microstructural scale. Highly coupled, multiphysics simulations are inherently computational expensive \cite{Zhang2019, Zhang2019a}, which is exacerbated by the need to perform them on volumes large enough to be considered representative.  To make matters worse, the inherent non-periodic nature of real tomographic data, combined with the typical use of ``mirror" boundary conditions means that regions near the edges of the simulated control volume will behave differently from those in the bulk. This leads to a further increase in the size of the simulated volume required, as the impact of the ``near edge" regions need to be eclipsed by the bulk response. Already common practice in the study of turbulent flow \cite{Lees1972, Henys2019}, the use of periodic boundaries enables much smaller volumes to be used, which can radically accelerate simulations. The flux maps shown in Figure \ref{Figure8_Periodic_boundaries} highlight the potential impact even for a simple diffusion simulation and the calculated transport parameters of these small volumes are much closer to the bulk response when periodic boundaries are implemented.\\

Examples of generated periodic (and similar non-periodic) volumes for both the Li-ion cathode and SOFC anode can be found in the supplementary materials accompanying this paper and the authors encourage the community to investigate their utility. A detailed exploration of the various methods for reconfiguring the generator’s architecture for the generation of periodic boundaries, as well as an analysis of the morphological and transport properties of the generated microstructures compared to the real ones are ongoing and will be presented in future work.\\

An additional benefit of the use of GANs in microstructural generation lies in the ability to interpolate in the continuous latent space to generate more samples of the same microstructure. The differentiable nature of GANs enables the latent space that parametrises the generator to be optimised. Li et al. \cite{Li2018} have implemented a Gaussian process to optimise the latent space in order to generate an optimum 2D two-phase microstructure. Other authors \cite{Yeh2017} have used an in-painting technique to imprint over the three-dimensional image some microstructural details that are only available in two-dimensional conditioning data. This process is performed by optimising the latent vector with respect to the mismatch between the observed data and the output of the generator \cite{Yeh2017, Mosser2018}. A potential implementation of the in-painting technique could involve adding information from electron backscatter diffraction (EBSD), such as crystallographic structure and orientation, into the already generated 3D structures, which would be of great interest to the electrochemical modelling community.\\

Future work will aim to extend the study by Li et al. \cite{Li2018} to perform an optimisation of the 3D three-phase microstructure based on desired morphological properties by optimising the latent space. One proposed pathway to improve these optimisation process would involve providing physical parameters to the GAN architecture. This could be achieved by adding a physics-specific loss component to penalise any deviation from a desired physical property \cite{Paganini2018}. It could also involve giving a physical meaning to the \textbf{z} space through the implementation of a Conditional GAN algorithm \cite{Isola2017}. With this, apart from the latent vector, the Generator has a second input \textbf{y} related to a physical property. Thus, the Generator becomes $G(\mathrm{\textbf{z}},\mathrm{\textbf{y}})$ and produces a realistic image with its corresponding physical property

\section{Conclusions}

This work presents a method for generating synthetic three-dimensional microstructures composed of any number of distinct material phases, through the implementation of DC-GANs. This method allows the model to represent the statistical and morphological properties of the real microstructure, which are captured in the weights of the trained discriminator and generator networks.\\

A pair of open-source, tomographically derived, three-phase microstructural datasets were investigated: a lithium-ion battery cathode and a solid oxide fuel cell anode. Various microstructural properties were calculated for 100 sub-volumes of the real data and these were compared to 100 instances of volumes created by the trained generator. The results showed excellent agreement across all metrics, although the synthetic structures showed a smaller variance compared to the training data, which is a commonly reported problem for DC-GANs and mitigation strategy will be reported in future work.\\

Two issues encountered when training the DC-GANs in this study were instability (likely due to a vanishing gradient) and moderate mode collapse. Both issues can be attributed to the GANs loss function and solutions have been suggested in the literature, the implementation of which will be explored in future work.\\

Two particular highlights of this work include the ability to generate arbitrarily large synthetic microstructural volumes and the generation of periodic boundaries, both of which are of high interest to the electrochemical modelling community. A detailed study of the impact of periodic boundaries on the reduction of simulation times is already underway.\\

Future work will take advantage of the continuity of the latent space, as well as the differentiable nature of GANs, to perform optimisation of certain morphological and electrochemical properties in order to discover improved electrode microstructures for batteries and fuel cells. 

\section*{Data Availability}
The study used open-access training data available from references listed in the manuscript. Many instances of the generated data are available at\url{https://github.com/agayonlombardo/threephase_porous_material}. All other generated data used is available from the authors on request.

\section*{Code Availability}
All the codes used in this manuscript can be accessed via the following link \url{https://github.com/agayonlombardo/threephase_porous_material}

\section*{Acknowledgements}
This work was supported by funding from both the CONACYT-SENER fund and the EPSRC Faraday Institution Multi-Scale Modelling project (\url{https://faraday.ac.uk/};  EP/S003053/1, grant number FIRG003). The Titan Xp GPU used for this research was kindly donated by the NVIDIA Corporation through their GPU Grant program. The authors would also like to thank various colleagues for their valuable comments: Prof. Stephen Neethling, Antonio Bertei, Tilman Roeder, Steven Kench, Vitaly Levdik, Donal Finegan and Harry Abernathy.  

\section*{Author Contributions}
A.G.L. wrote the code. S.J.C. supervised the project. S.J.C. and A.G.L. conceived the idea and performed the analysis. L.M. provided the theoretical foundations and expertise in GANs. N.P.B. offered supervision and guidance on the wider impact of the work. A.L.G. and S.J.C. wrote the manuscript.

\section*{Competing interests}
The authors declare no competing interests.
\section*{References}
\addcontentsline{toc}{section}{References}
\begingroup
\def\addvspace#1{}

	\renewcommand{\refname}{ \vspace{-\baselineskip}\vspace{-1.1mm} }
	\bibliographystyle{abbrv}
	\bibliography{AGL_GANs}
\endgroup

\end{document}


\title{Pores for thought: The use of generative adversarial networks for the stochastic reconstruction of 3D multi-phase electrode microstructures with periodic boundaries\\
	Supplementary information}
	
	\author[1]{Andrea Gayon-Lombardo\thanks{a.gayon-lombardo17@imperial.ac.uk}}
	\author[2]{Lukas Mosser}
	\author[1]{Nigel P. Brandon\thanks{n.brandon@imperial.ac.uk}}
	\author[3]{Samuel J. Cooper\thanks{samuel.cooper@imperial.ac.uk}}
	\affil[1]{{\textit{\footnotesize Department of Earth Science and Engineering, Imperial College London, London SW7 2BP}}}
	\affil[2]{{\textit{\footnotesize Earth Science Analytics ASA, Professor Olav Hanssens vei 7A, 4021 Stavanger, Norge}}}
	\affil[3]{{\textit{\footnotesize Dyson School of Design Engineering, Imperial College London, London SW7 2DB}}}
	
	\maketitle
	
	\section{Introduction to GANs}
	The idea of GANs originated from game theory, as a two-player game between the generator and the discriminator. The generator, ($G(\mathrm{\textbf{z}})$), takes as input a latent vector z and creates samples ($\mathrm{\textbf{x}}_G$) that are intended to approximate the real data ($\mathrm{\textbf{x}}_\mathrm{data}$). The discriminator, $D(\mathrm{\textbf{x}})$, tries to distinguish the generated samples from the real ones. This concept was memorably brought to life by Goodfellow et al. \cite{Goodfellow2014} who frame the generator as a counterfeiter producing fake currency, while the discriminator is a police detective who aims to distinguish the fake currency from the real. In this two-player game, both continuously improve by pushing the skills of the other.\\
	 
	The generator and the discriminator are differentiable functions with respect to their inputs and parameters. The generator, defined as $G_{\theta^{(G)}} (\mathrm{\textbf{z}})$, maps random variables from the latent space $\mathrm{\textbf{z}}$ into the image domain, and uses $\theta^{(G)}$ as parameters. The latent space \textbf{z} is defined by a prior distribution $p_\mathrm{z}$, corresponding to a normal distribution $\mathcal{N}(0,1)$; $\mathrm{\textbf{z}}$ is a tensor of dimensionality d, composed of independent, normally distributed, real variables:
	
	\begin{equation}
	\mathrm{\textbf{z}} \sim \mathcal{N}(0,1)^{d \times 1 \times 1 \times 1}
	\end{equation}
	\begin{equation}
	G_{\theta^{(G)}}:\mathrm{\textbf{z}} \rightarrow \mathbb{R}^{3 \times 64 \times 64 \times 64}
	\end{equation}
	
	The discriminator, defined by its parameters $\theta^{(D)}$, as $D_{\theta^{(D)}} (\mathrm{\textbf{x}})$ receives samples from the real data $\mathrm{\textbf{x}}_\mathrm{data}$ and from the synthetic data generated by the generator $G(\mathrm{\textbf{z}})$:
	
	\begin{equation}
	\mathrm{\textbf{x}} \sim \mathbb{R}^{3 \times 64 \times 64 \times 64}
	\end{equation}
	\begin{equation}
	D_{\theta^{(D)}}:\mathrm{\textbf{x}} \rightarrow [0,1]
	\end{equation}
	
	Through this adversarial game, the two players are in constant competition, where the discriminator attempts to label the real and fake data correctly (\textit{i.e.} $D(\mathrm{\textbf{x}}_\mathrm{data}) = 1$ and $D(G(\mathrm{\textbf{z}})) = 0$), while the generator tries to fool the discriminator into misclassifying synthetic data as real (\textit{i.e.} $D(G(\mathrm{\textbf{z}})) = 1$). This problem is formally defined as a minimax game between the generator and discriminator where the objective function to optimise is $V(G,D)$ \cite{Goodfellow2014}:  
	
	\begin{equation}
	\min_G\left(\max_D \left(V(G,D) \right)\right)  = \mathbb{E}_{\mathrm{\textbf{x}}\sim p_{\mathrm{data}}(\mathrm{\textbf{x}})} \left[\log \left(D(\mathrm{\textbf{x}})\right) \right] + \mathbb{E}_{\mathrm{\textbf{z}}\sim p_{\mathrm{z}}(\mathrm{\textbf{z}})} \left[\log \left(1 - D\left(G(\mathrm{\textbf{z}})\right) \right) \right]
	\end{equation}
	
	\section{Sample details}
	\subsection{Li-ion cathode}
	Li-ion cathode
	This material was generated at the Cell Analysis, Modelling and Prototyping (CAMP) facility at the Argonne National Laboratory. The electrode is composed of three phases: the active material corresponding to $\mathrm{Li}(\mathrm{Ni}_{0.5}\mathrm{Mn}_{0.3}\mathrm{Co}_{0.2})\mathrm{O}_2$ (NMC532, TODA America Inc.), the carbon/binder domain composed of C45 carbon (Timcal) and polyvinylidene fluoride (PVDF) (Solvay, Solef 5130), and the porous phase. Since the carbon/binder domain is not visible directly at the X-ray computed tomography (CT) images, the authors generated this phase over the three-dimensional geometry of the X-CT images \cite{Usseglio-Viretta2018}.\\
	
	\subsection{SOFC anode}
	This material consists of a commercial anode-supported button MSRI (Materials and Systems Research, Inc.) cell composed of three phases: Yttria-stabilised zirconia (YSZ), nickel (Ni) and the porous phase. The total size of the electrode sample is $\approx 4 \times 10^8$ voxels ($1900 \times 1697 \times 124$ voxels). \cite{Hsu2018} This voxelised image is cropped into a total of 45,492 overlapping sub-volumes of $64^3$ voxels each, at a spacing of 16 to be used as training database for reconstruction. This image size is sufficient enough to represent the TPCF of the whole electrode sample.\\
	
	\section{One-hot encoding}
\begin{figure}[]
		\centering
		\includegraphics[clip, trim= 4cm 4cm 4cm 2cm, width=1\textwidth]{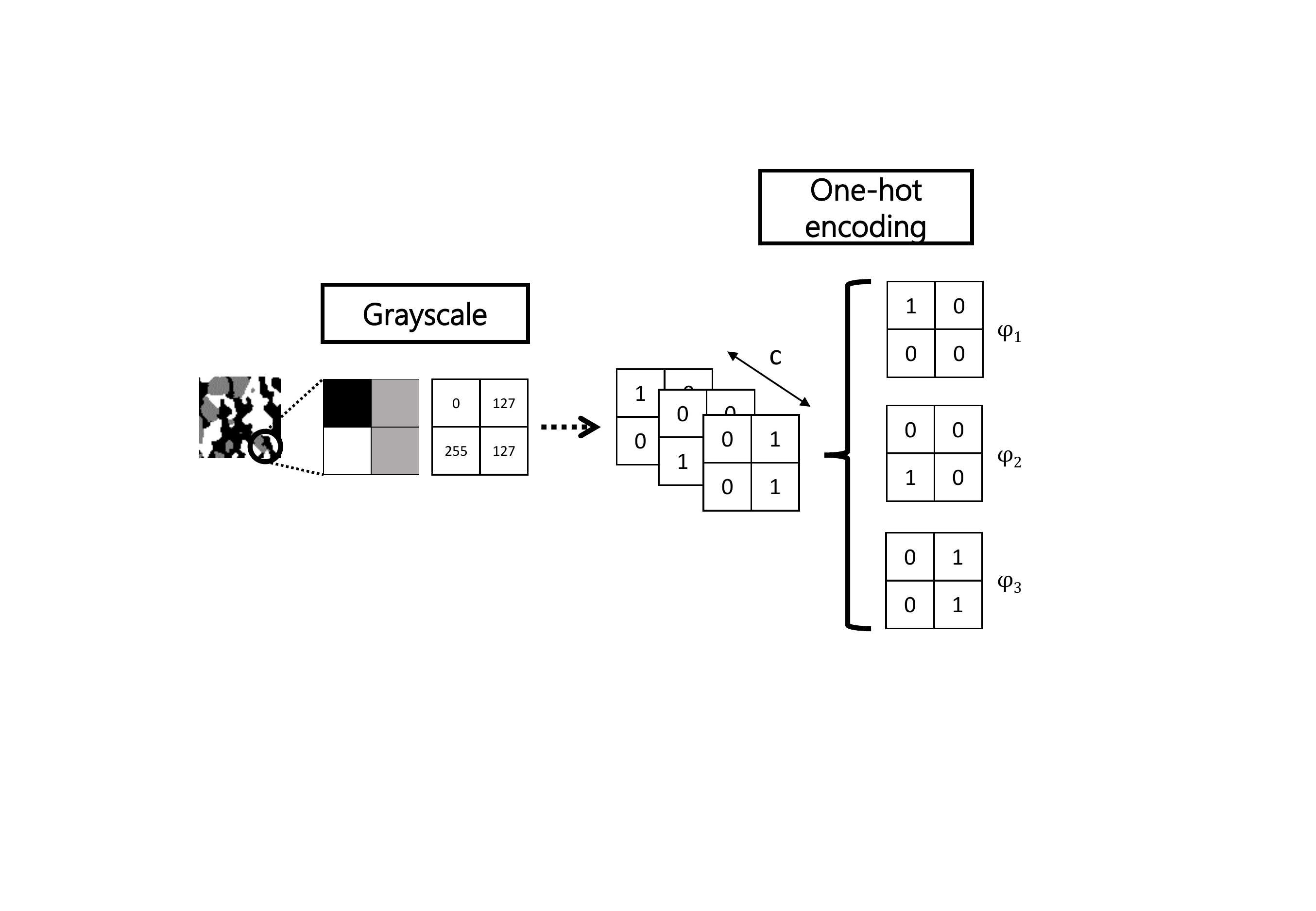}
		\caption[]
		{{\footnotesize One-hot encoding process}}    
		\label{Figure1_One_hot}
	\end{figure}

	A visual representation of the one-hot encoding method is shown in figure \ref{Figure1_One_hot}.
	
	\section{Microstructural parameters}
	\subsection{Morphological parameters}
	The phase volume fraction $\phi_i$ is defined as:
	
	\begin{equation}
	\phi_i = \frac{V_i}{V}
	\end{equation}
	
	where $V_i$ refers to the volume of phase $i$, calculated as the percentage of voxels corresponding to this phase, and $V$ to the total volume of the microstructure domain.\\
	
	The specific surface area $S_V$ is defined as:
	
	\begin{equation}
	S_{V_i} = \frac{S_{A_i}}{V} = \frac{1}{V} \int d S_{A_i}
	\end{equation}
	
	where $S_{A_i}$ refers to the total surface area of the interface between phase $i$ and the rest of the phases.\\
	
	The TPB density is defined as the length of the intersection among three phases $i$, $j$, and $k$, normalised by the total volume of the microstructure domain. For a cuboid lattice, a TPB is defined as the length of the edges where three of the four connecting voxels contain different phases \cite{Cooper2016}.
	
	\subsection{Relative diffusivities}
	The tortuosity factor $\tau$ is defined by Cooper et al. \cite{Cooper2016} as the resistance offered by the geometry of the porous medium towards diffusive transport. Other authors define the tortuosity factor or geometric tortuosity $\tau^{\mathrm{geo}}$ as the ration of the effective pathway length $l^{\mathrm{eff}}$ over the length of the sample $l$. In both definitions this parameter measures the geometric deviation of the conductive phase from the straight path. Two properties of the tortuosity factor are presented by Cooper et al. \cite{Cooper2016}: firstly, when $\tau\,=\,1$, the flow must be direct, and secondly ofr al systems $\tau\, \geq\, 1$. The effect of the tortuosity factor is inversely related to the diffusivity, as expressed in equation \ref{eq_diff}:
	
	\begin{equation}
	D^{\mathrm{eff}} = 	D^{\mathrm{o}} \frac{\phi_i}{\tau}
	\label{eq_diff}
	\end{equation}
	
	where $\phi_i$ is the volume fraction of the conductive phase $D^{\mathrm{o}}$ is the intrinsic diffusivity of the conductive phase, and $D^{\mathrm{eff}}$ is the effective diffusivity through the conductive phase. The tortuosity factor is obtained as the ratio between the steady-state diffusive flow through a conductive phase, $F_\mathrm{p}$, and a fully dense volume of the same size, $F_\mathrm{CV}$ \cite{Cooper2016}.
	
	\begin{equation}
	F_\mathrm{p} = -A_\mathrm{CV} D \frac{\phi_i}{\tau} \frac{\Delta C}{L_\mathrm{CV}}
	\end{equation}
	
	\begin{equation}
	F_\mathrm{CV} = -A_\mathrm{CV} D \frac{\Delta C}{L_\mathrm{CV}}
	\end{equation}
	
	where $D$ is the diffusivity of the conductive phase, $C$ is the local concentration of the diffusing species, $A_\mathrm{CV}$ is the cross-sectional area of the control volume and $L_\mathrm{CV}$ is the length of the control volume \cite{Cooper2016}.
	
	\newpage
	\section{Relative diffusivity extended results}
	\subsection{Lithium-ion cathode}
\begin{figure}[ht]
	\centering
	\includegraphics[clip, trim= 3cm 0cm 2.5cm 0cm, width=1\textwidth]{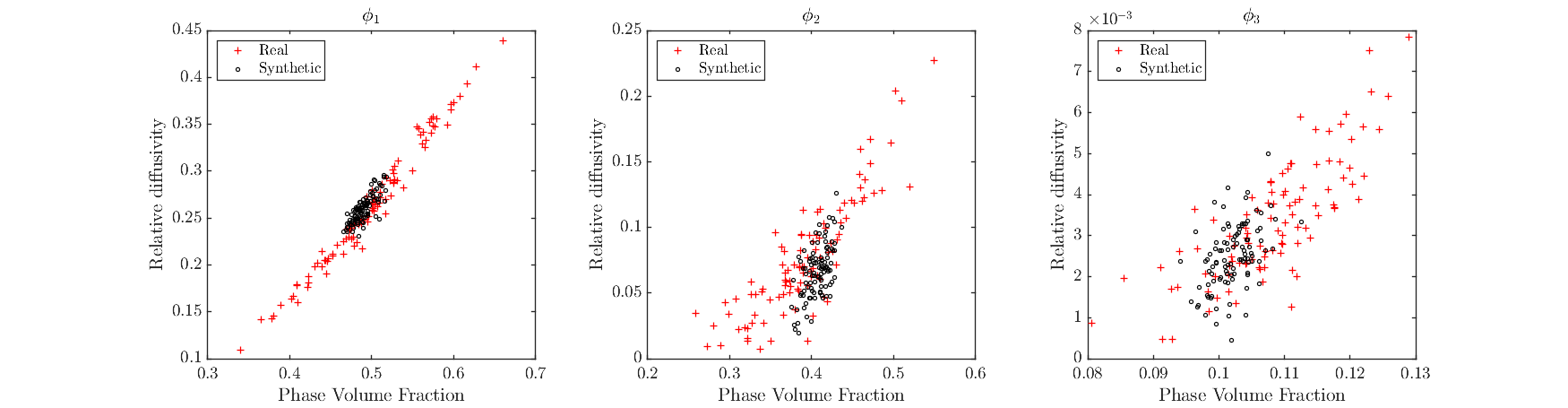}
	\caption[]
	{{\footnotesize Relative diffusivity w.r.t. phase volume fraction for the three phases for the Li-ion cathode}}    
	\label{Figure2_Liion}
	\end{figure}

	A representation of the relative diffusivity w.r.t the volume fraction of the three phases for the Li-ion cathode is shown in figure \ref{Figure2_Liion}.

	\subsection{SOFC anode}
\begin{figure}[ht]
		\centering
		\includegraphics[clip, trim= 3cm 0cm 2.5cm 0cm, width=1\textwidth]{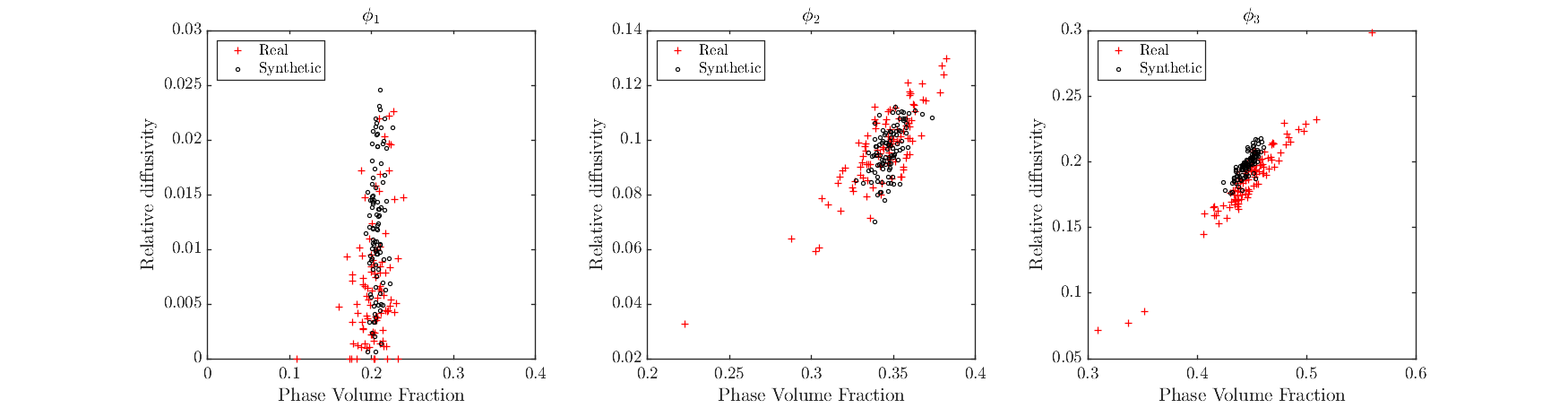}
		\caption[]
		{{\footnotesize Relative diffusivity w.r.t. phase volume fraction for the three phases for the SOFC anode}}    
		\label{Figure2_SOFC}
	\end{figure}

	A representation of the relative diffusivity w.r.t the volume fraction of the three phases for the SOFC anode is shown in figure \ref{Figure2_SOFC}.
	
	\newpage
	\section{TPCF extended results}
	\subsection{Lithium ion cathode}
\begin{figure}[ht]
	\centering
	\includegraphics[clip, trim= 1cm 5cm 1cm 5cm, width=1\textwidth]{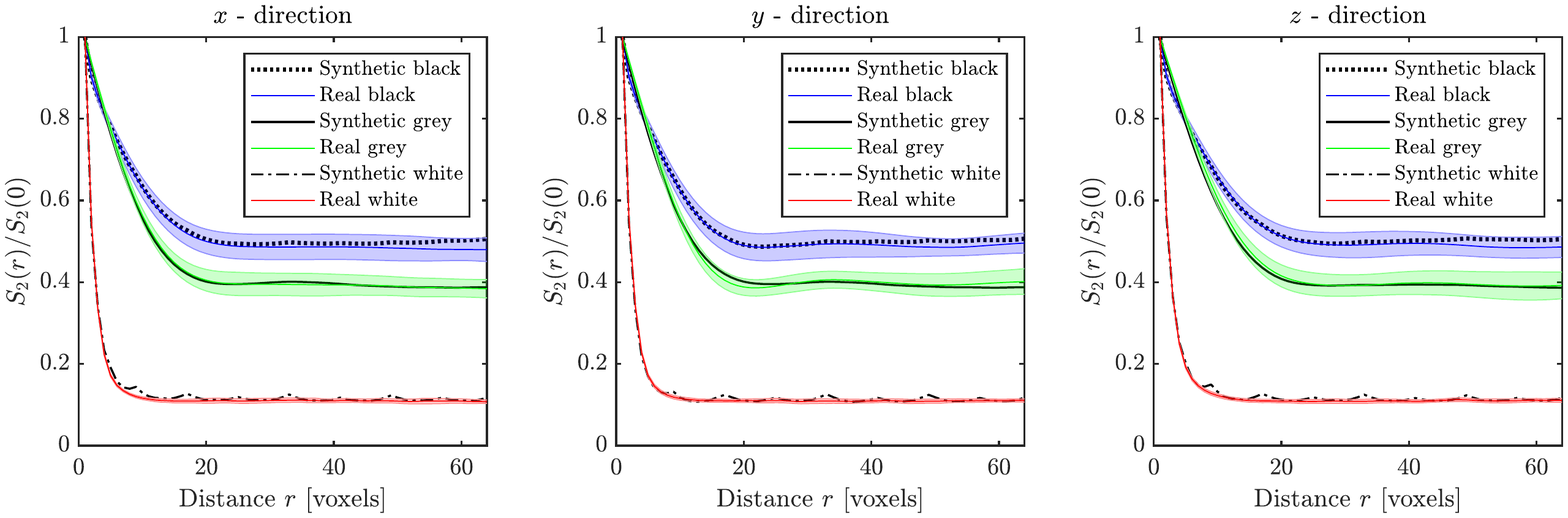}
	\caption[]
	{{\footnotesize $S_2(\mathrm{\textbf{r}}) / S_2(0)$ of the black phase corresponding to the pores (blue line), grey phase corresponding to the NMC particles (green line) and white phase corresponding to the binder (red line)   in a Li-ion cathode.}}    
	\label{FigureS3_TPCF_Liion}
	\end{figure}

	Figure \ref{FigureS3_TPCF_Liion} shows the results of the normalised TPCF (\textit{i.e.} $S_2(\mathrm{\textbf{r}}) / S_2(0)$) of the three phases (pores, NMC-532 and binder) along the three directions. For all cases, the synthetic data follows the same trend as the real dataset, and is able to represent the anisotropy of some phases in the material. Table \ref{Table_Liion decay} shows the results of a detailed description of the decay along the three axis and the stabilisation of $S_2(\mathrm{\textbf{r}}) / S_2(0)$ for each phase.\\

	\begin{table}[ht]
		\caption{Detailed description of the normalised TPCF for the Li-ion cathode along the three directions for each of the three phases (pores, NMC and binder)}
		\resizebox{\textwidth}{!}{
		\begin{tabular}{c|c|c|c|c}
			\hline
			\textbf{Phase} & \textbf{x} & \textbf{y} & \textbf{z} & \textbf{Stabilisation} \\
			\hline
			Black (pores) & near exponential decay & near exponential decay & near exponential decay & 0.5 \\
			Grey (NMC) & near exponential decay & small hole effect & near exponential decay & 0.39 \\
			White (binder) & exponential decay & exponential decay & exponential decay & 0.11 \\
			\hline
		\end{tabular}}
		\label{Table_Liion decay}
	\end{table}
	
	\newpage
	\subsection{SOFC anode}
\begin{figure}[ht]
	\centering
	\includegraphics[clip, trim= 1cm 5cm 1cm 5cm, width=1\textwidth]{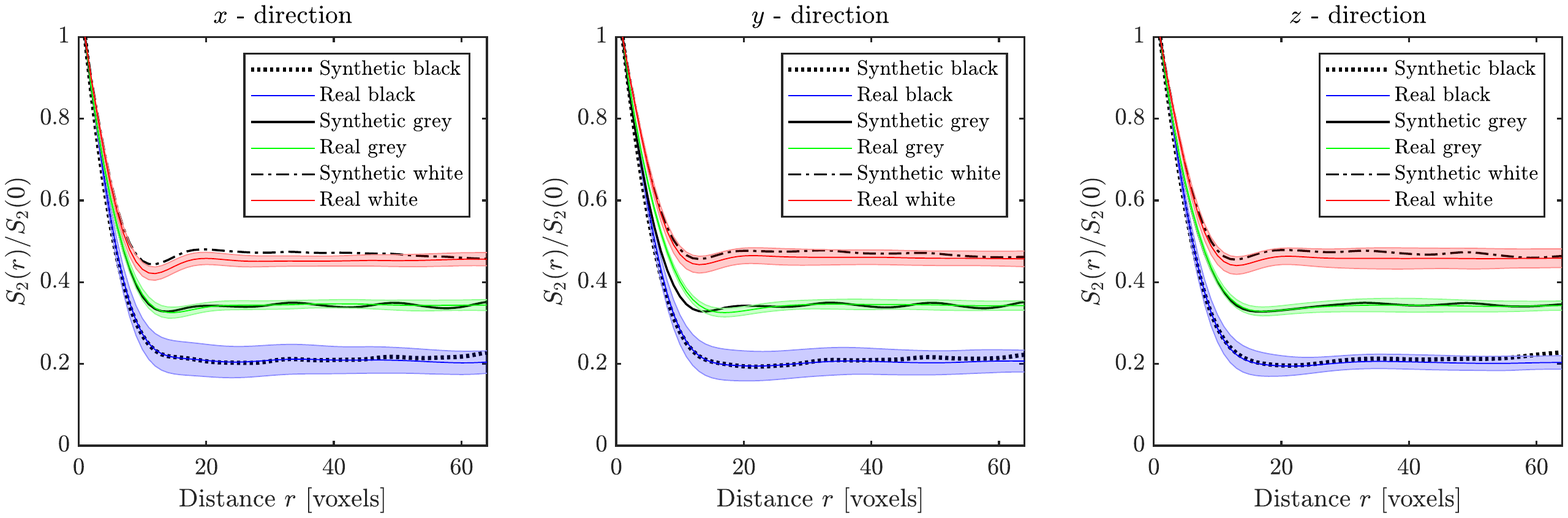}
	\caption[]
	{{\footnotesize $S_2(\mathrm{\textbf{r}}) / S_2(0)$ of the black phase corresponding to the pores (blue line), grey phase corresponding to the Nickel (green line) and white phase corresponding to the YSZ (red line)   in a LI-ion cathode.}}    
	\label{FigureS4_TPCF_SOFC}
	\end{figure}

	Figure \ref{FigureS4_TPCF_SOFC} shows the results of the normalised TPCF (\textit{i.e.} $S_2(\mathrm{\textbf{r}}) / S_2(0)$) of the three phases (pores, Ni and YSZ) along the three directions. For all cases, the synthetic data follows the same trend as the real dataset, and is able to represent the anisotropy of some phases in the material. Table \ref{Table_SOFC decay} shows the results of a detailed description of the decay along the three axis and the stabilisation of $S_2(\mathrm{\textbf{r}}) / S_2(0)$ for each phase.\\

	\begin{table}[ht]
	\caption{Detailed description of the normalised TPCF for the SOFC anode along the three directions for each of the three phases (pores, Ni and YSZ)}
	\resizebox{\textwidth}{!}{
		\begin{tabular}{c|c|c|c|c}
			\hline
			\textbf{Phase} & \textbf{x} & \textbf{y} & \textbf{z} & \textbf{Stabilisation} \\
			\hline
			Black (pores) & near exponential decay & near exponential decay & near exponential decay & 0.2 \\
			Grey (Ni) & small hole effect & small hole effect & near exponential decay & 0.35 \\
			White (YSZ) & pronounced hole effect & pronounced hole effect & pronounced hole effect & 0.45 \\
			\hline
	\end{tabular}}
	\label{Table_SOFC decay}
	\end{table}

	\newpage
	\section{Representativity}
	This method uses a set of sub-volumes obtained from X-ray tomographic images for the Li-ion cathode and PFIB-SEM for the SOFC anode, as training sets for the GANs. In this work, the methodology was applied for the reconstruction of two different microstructures commonly used for energy storage technologies: a Li-ion cathode and an SOFC anode. The size of the training images was selected according to the size of the structuring element of the microstructure. For the Li-ion cathode, there is a clear characteristic feature consisting on the NMC particle size (\textit{i.e.} grey phase). The average particle diameter is 35 voxels, so a $64^3$ voxels sub-volume can capture the volume of an entire particle and most of the volume of neighbouring particles. In the case of the SOFC anode, all three phases form continuous networks so no structuring element (such as grain size) \cite{Mosser2017} is observed. Therefore, a standard training size of $64^3$ voxels \cite{Radford2015} was implemented based on a level of representativity in the measured volume fraction.\\
	
	The image training size is a topic of interest as it is closely related to the representativeness of the training set. As previously shown, in the case of structures with a characteristic feature such as grain size or particle size, it is possible to use this as a criteria for selecting a training image size; however for the cases where the structure is irregular (as is the case of the SOFC anode), highly porous or a continuously fibrous media, choosing the ideal training size becomes a challenging task.\\
	
	It is well established among authors \cite{Lu1990, Yeong1998, Yeong1998a, Manwart2000, Sheehan2001} that a two-point statistics function provides an accurate description of the long-range properties and therefore could provide an insight into the size of a representative elementary volume (REV). In some cases, the size of a representative volume can be estimated based on the stabilisation value of the two-point statistic function $S_2(\mathrm{\textbf{r}})$ \cite{Mosser2018}. This would require an evaluation of $S_2(\mathrm{\textbf{r}})$ along the three directions and in all phases present in the microstructure. In the two materials analysed in this work, stabilisation was achieved at an average of 40 voxels along the three directions in the three phases. This shows that a selection of $64^3$ sub-volume as training set provides a fair representation of the microstrucutre, at least in terms of volume fraction.\\
	
	Although the $S_2(\mathrm{\textbf{r}})$ provides essential information about the constitution and long-range properties of microstructures, for more complex samples where the $S_2(\mathrm{\textbf{r}})$ does not achieve stabilisation, a better understanding of the REV in terms of the microstructural characterisation parameters is required \cite{Mosser2018}. As mentioned by Mosser et al. \cite{Mosser2017}, an estimation of the REV of the specific surface area is more representative of the morphology of a porous medium. In the case of three-phase data, an analysis of a REV of the TPB could provide an even deeper insight of the representativeness of the sub-volume since this particular parameter accounts for the interaction of the three phases. To the authors’ knowledge no conclusive research has been reported in the literature which provides a full analysis on the representativeness of properties like specific surface area and TPB. An initial approximation would involve a statistical analysis of such properties obtained experimentally from sub-volumes of tomographic data at different sections of a larger volume. This approach however is exhaustive in terms of experimentation, and unique for each microstructure, therefore it cannot be generalised for a wide range of porous materials.  Based on this, a thorough research on the implementation of long range functions, such as the two-point statistics function, that can provide information on the REV of additional properties is recommended as future work.
	
	\section{Generating larger volumes}
	The minimum generated samples are the same size as the training data sub-volumes (i.e. 643 for both cases analysed in this work), but can be increased to any arbitrarily large size by increasing the size of the input z.\\
	
\begin{figure}[ht]
	\centering
	\includegraphics[clip, trim= 2cm 10cm 12cm 1cm, width=1\textwidth]{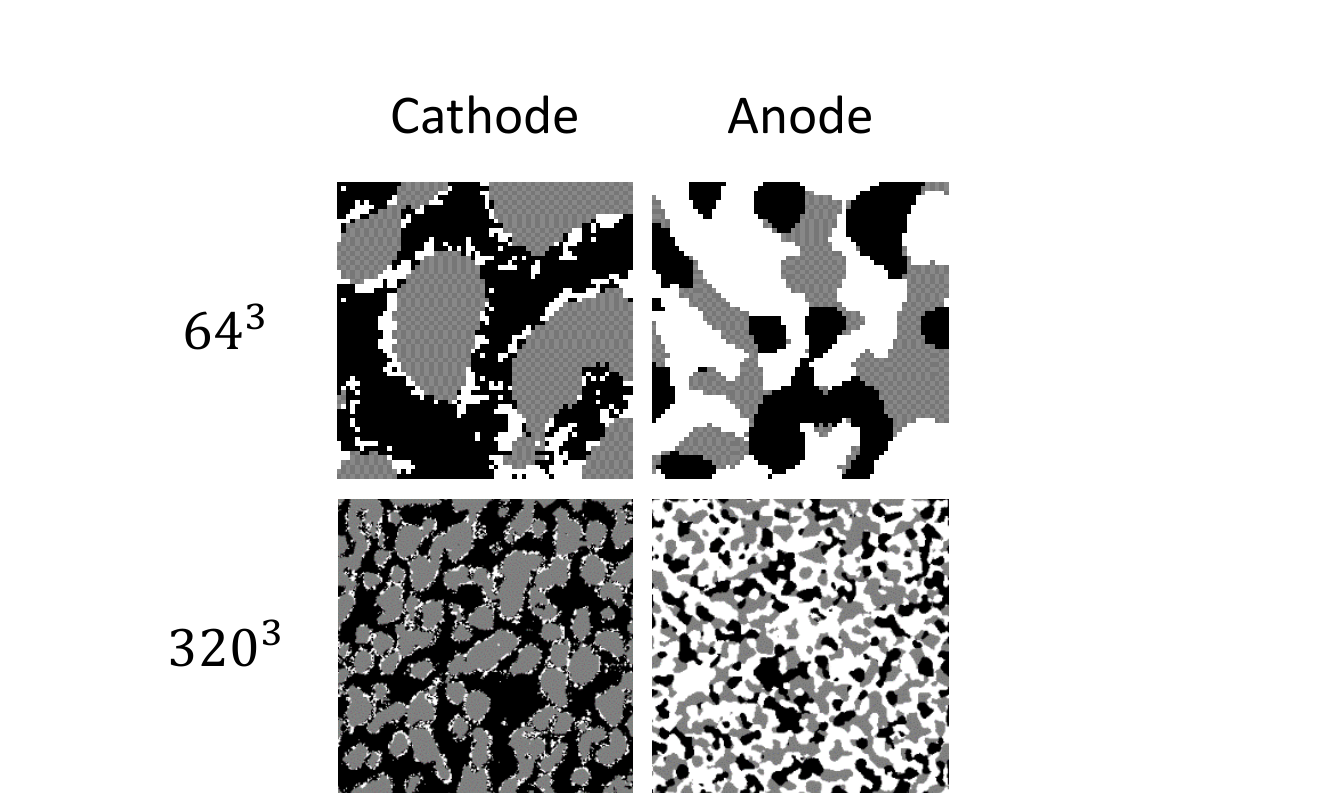}
	\caption[]
	{{\footnotesize Large generated images of SOFC anode and Li-ion cathode based on previously trained generator with training data of $64^3$ voxels.}}    
	\label{FigureS5_Large_images}
	\end{figure}

	In this work samples as large as 3203 voxels were generated. The generated images are shown in figure \ref{FigureS5_Large_images}
	
	\newpage
	\section{Computational expense}
\begin{figure}[ht]
		\centering
		\includegraphics[clip, trim= 6cm 4.5cm 6cm 4.5cm, width=1\textwidth]{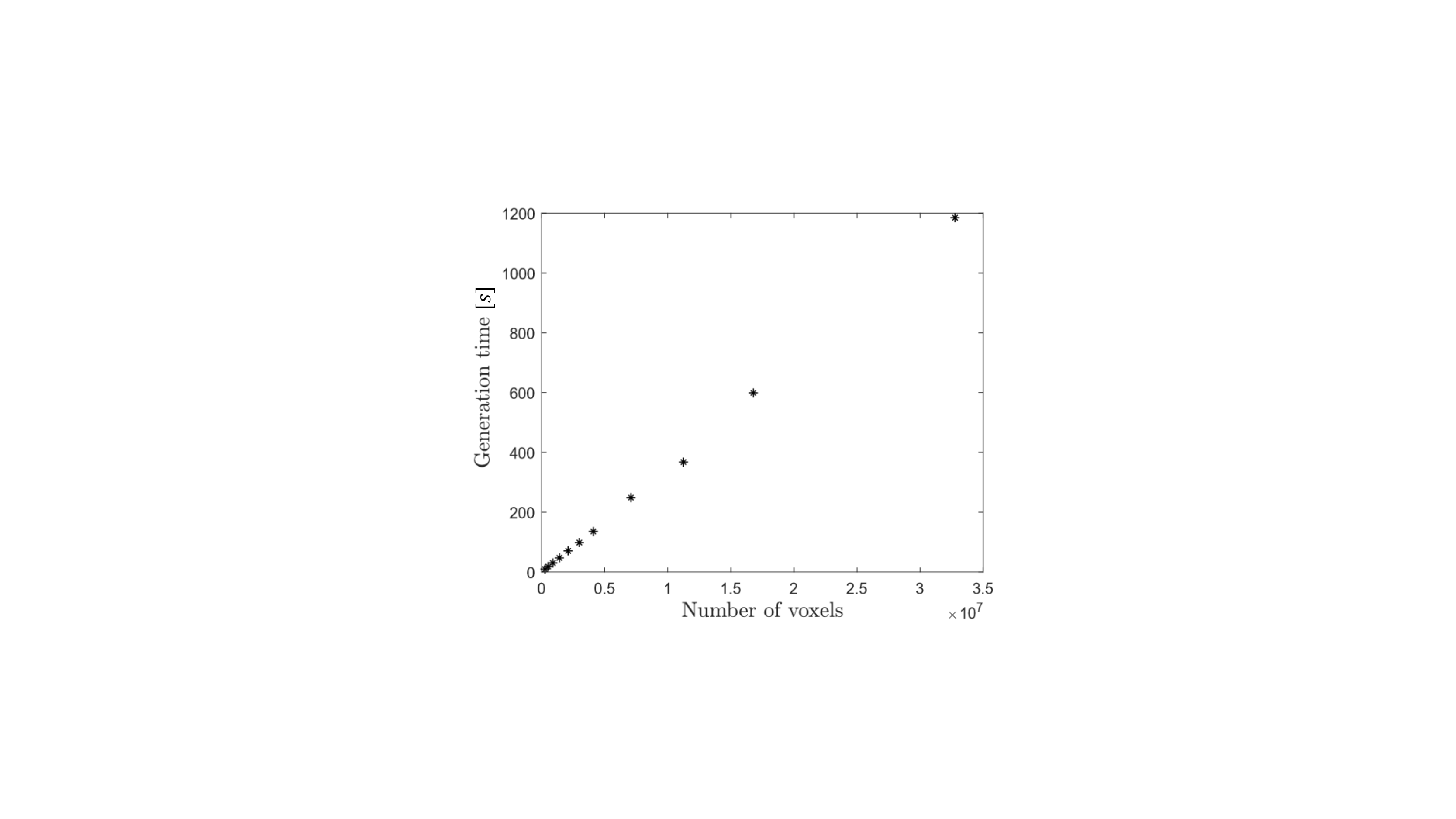}
		\caption[]
		{{\footnotesize Generation time w.r.t number of voxels}}    
		\label{FigureS6_Generation_time}
	\end{figure}

	GANs present the capability of generating larger images with the same probability distribution function of the real dataset by increasing the size of the z space and using this larger latent space as input of the trained generator. The size of the larger images is given by the following equation:

	\begin{equation}
	\mathrm{\textbf{z}} \sim \mathcal{N}(0,1)^{100 \times 1*\alpha \times 1*\alpha \times 1*\alpha}
	\end{equation}
	\begin{equation}
	G_{\theta}(\mathrm{\textbf{z}}) \rightarrow \mathbb{R}^{3 \times \lambda \times \lambda \times \lambda}
	\end{equation}
	\begin{equation}
	\lambda = 64 + (\alpha-1) \times 16
	\end{equation}
	
	where $\alpha$ corresponds to the incremental factor. Since the generation of larger images does not require further training, the generation is comparatively faster than the training process. The increase in computational time with respect to an increase in the size of the generated images is linear, as shown in figure \ref{FigureS6_Generation_time}. This linear trend is in agreement with the results presented by Mosser et al. \cite{Mosser2017, Mosser2018}.\\
	
	\newpage
	\section{Mode collapse}
	In the presence of mode collapse, the generator does not have an incentive to move to a different area within the optimisation space since it has found a local minimum. There are different levels of mode collapse, the perfect case being when the generator  memorises the training set, thus producing the same image every time. A thorough characterisation of the microstructure according to their morphological features can determine the accuracy of the generator in representing the real data; however, it does not account for the existence or level of mode collapse. An approach suggested by Radford et al. \cite{Radford2015} is implemented in this work to visually analyse the existence of memorisation of the dataset. This approach consists on performing an interpolation between two points in the latent space z, given by equations \ref{eq_int_1} and \ref{eq_int_2}:

	\begin{equation}
	\mathrm{\textbf{z}_{start}},\ \mathrm{\textbf{z}_{end}} \sim \mathcal{N}(0,1)^{100 \times 1 \times 1 \times 1},\ \beta \in [0,1]
	\label{eq_int_1}
	\end{equation}
	\begin{equation}
	\mathrm{\textbf{z}_{int}} = \beta * \mathrm{\textbf{z}_{start}}	+ (1-\beta) *\mathrm{\textbf{z}_{end}}
	\label{eq_int_2}
	\end{equation}
	
\begin{figure}[ht]
		\centering
		\includegraphics[clip, trim= 4cm 10.5cm 4cm 2cm, width=1\textwidth]{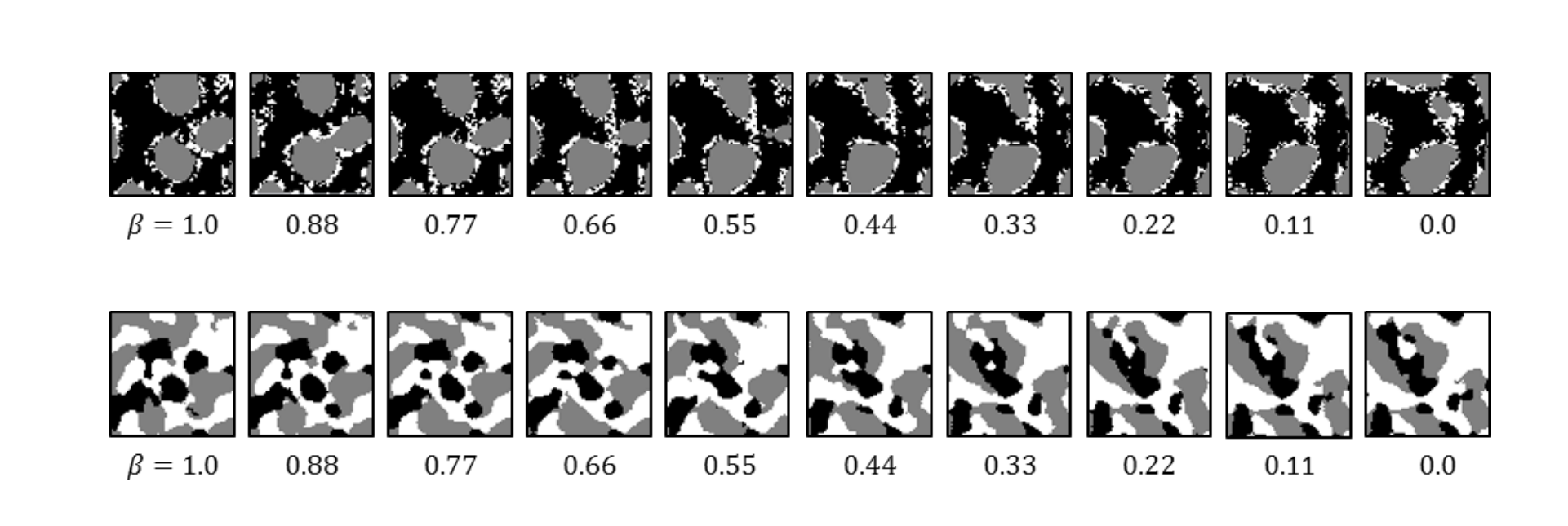}
		\caption[]
		{{\footnotesize Interpolation in the latent space z from $\mathrm{\textbf{z}_{start}}$ to $\mathrm{\textbf{z}_{end}}$. The smooth transition between both points in both microstructures shows the inexistence of a total mode collapse.}}    
		\label{FigureS7_Interpolation}
	\end{figure}

	Where $\beta$ is a set of integers between 0 and 1. The smooth transition in the images generated between the two points $G(\mathrm{\textbf{z}_{start}})$ and $G(\mathrm{\textbf{z}_{end}})$ indicates that the generator has not memorised the training set but has accurately learned a lower-dimensional representation of \textbf{z}. \cite{Mosser2018} The interpolation between two points in the \textbf{z} space is shown in figure \ref{FigureS7_Interpolation}.\\
	
	\newpage
	\section{Uncertainty maps}
\begin{figure}[ht]
		\centering
		\includegraphics[clip, trim= 0cm 6cm 4cm 0cm, width=1\textwidth]{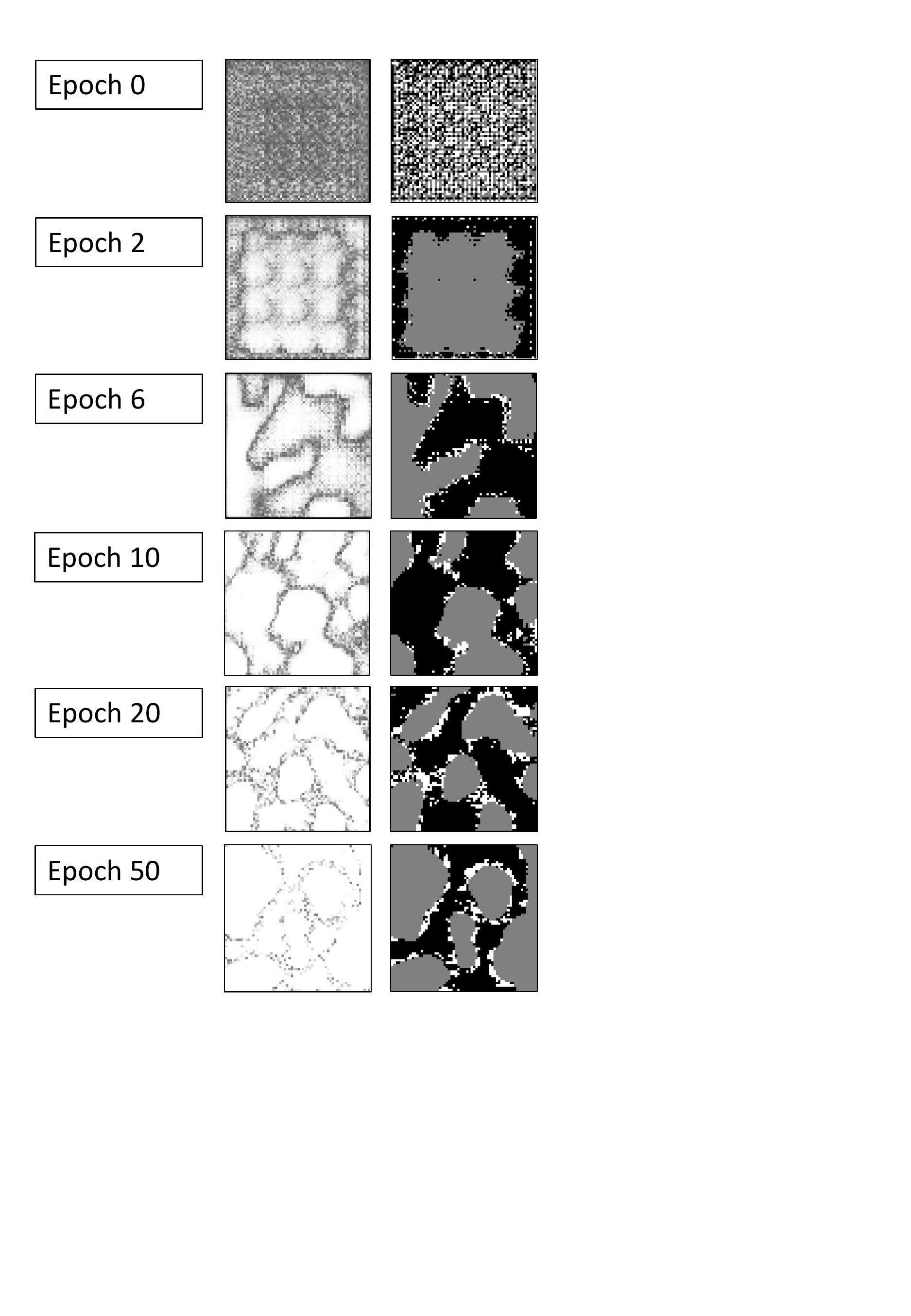}
		\caption[]
		{{\footnotesize Uncertainty maps along the training process for Li-ion cathode.}}    
		\label{FigureS8_Uncertainty_Liion}
	\end{figure}

\begin{figure}[ht]
	\centering
	\includegraphics[clip, trim= 0cm 6cm 4cm 0cm, width=1\textwidth]{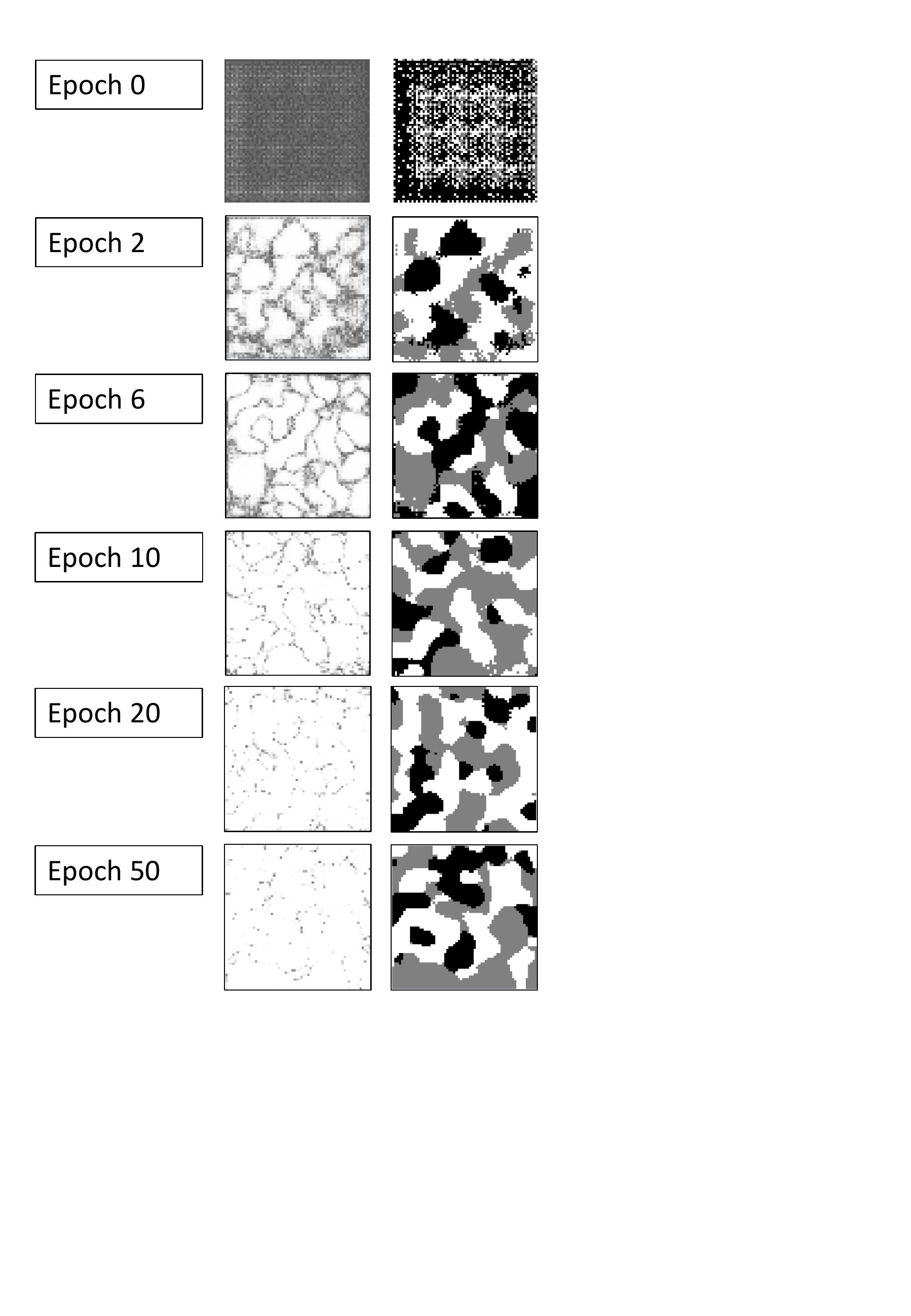}
	\caption[]
	{{\footnotesize Uncertainty maps along the training process for SOFC anode.}}    
	\label{FigureS9_Uncertainty_SOFC}
\end{figure}

	Figures \ref{FigureS8_Uncertainty_Liion} and \ref{FigureS9_Uncertainty_SOFC} show the maps of uncertainty for the generated images of the Li-ion cathode and SOFC anode respectively, as a function of the number of epochs during the training process. The grey-scale maps show the ``certainty” with which the generator assigns a phase ($\phi_i$) to each voxel. A high certainty is represented as a white voxel, while a low certainty is represented as a black voxel. It is seen that a higher uncertainty exists at the interphases, while low uncertainty exists at the bulk. These uncertainty decreases throughout the training process until epoch 50 where most of the uncertainty map is white.\\
	
\newpage
\clearpage
\section*{References}
\addcontentsline{toc}{section}{References}
\begingroup
\def\addvspace#1{}

\renewcommand{\refname}{ \vspace{-\baselineskip}\vspace{-1.1mm} }
\bibliographystyle{abbrv}
\bibliography{Sup_information}
\endgroup